\documentclass{article} 
\usepackage[preprint]{colm2026_conference}

\usepackage{microtype}
\usepackage{hyperref}
\usepackage{url}
\usepackage{booktabs}
\usepackage{amsfonts}       
\usepackage{nicefrac}       
\usepackage{float}          
\usepackage{amsmath}
\usepackage{graphicx}
\usepackage{bm, upgreek}
\usepackage{amssymb}
\usepackage{array}
\usepackage{multirow}
\usepackage{mathtools}
\usepackage{arydshln}
\usepackage{tabularx}
\usepackage{color}
\usepackage{soul}
\usepackage{caption}
\usepackage{xcolor}
\usepackage{comment}
\usepackage{adjustbox}
\usepackage{cleveref}
\usepackage[ruled,vlined]{algorithm2e}
\usepackage{makecell}
\usepackage{diagbox}
\usepackage{subfigure}
\usepackage{subcaption}
\usepackage{xcolor, soul}
\usepackage{wrapfig}
\usepackage{pifont}
\usepackage[most]{tcolorbox}
\definecolor{bluecolor}{HTML}{0000FF}
\definecolor{greencolor}{HTML}{8CD0A4}
\definecolor{yellowcolor}{HTML}{F9D17C}
\definecolor{redcolor}{HTML}{FF0000}

\definecolor{black}{rgb}{0,0,0}

\newcommand{\goodmetric}[1]{\textbf{\textcolor{green!70!black}{#1}}} 
\newcommand{\badmetric}[1]{\textbf{\textcolor{red!90!black}{#1}}}
\usepackage{scalerel,xparse}

\RestyleAlgo{ruled}
\newlength{\commentWidth}
\SetCommentSty{mycommfont}

\usepackage{lineno}

\definecolor{darkblue}{rgb}{0, 0, 0.5}
\hypersetup{colorlinks=true, citecolor=darkblue, linkcolor=darkblue, urlcolor=darkblue}

\title{Thinking Sparks!: Emergent Attention Heads in Reasoning Models During Post Training}

\author{Yein Park$^{1,3}$, Minbyul Jeong$^{2*}$, Jaewoo Kang$^{1,3}$\thanks{Corresponding authors} \\
  Korea University$^1$ \quad
  Upstage AI$^2$ \quad
  AIGEN Sciences$^3$ \\
  \{522yein, kangj\}@korea.ac.kr \quad
  minstar@upstage.ai
}

\begin{document}

\ifcolmsubmission
\linenumbers
\fi

\maketitle

\begin{abstract}
The remarkable capabilities of modern large reasoning models are largely unlocked through post-training techniques such as supervised fine-tuning (SFT) and reinforcement learning (RL).
However, the architectural mechanisms behind such improvements remain largely opaque.
In this work, we use circuit analysis to demonstrate that post-training for complex reasoning sparks the emergence of novel, functionally specialized attention heads.
These heads collectively support structured reasoning and computation.
Our comparative analysis across various model families reveals that these emergent heads evolve differently under different training regimes.
Distillation and SFT foster a cumulative addition of stable reasoning heads. 
In contrast, group relative policy optimization (GRPO) operates in a dynamic search mode: relatively few attention heads are iteratively activated, evaluated, and pruned, with their survival closely tracking fluctuations in the task reward signal.
Furthermore, we find that controllable ``think on/off" models do not possess dedicated ``thinking" heads.
Instead, turning off explicit reasoning triggers a broader—but less efficient—set of compensatory heads.
Through ablation and qualitative analyses, we connect these circuit-level dynamics to a crucial performance trade-off:
strengthened heads enable sophisticated problem-solving strategies for difficult problems but can also introduce ``over-thinking" failure modes, such as calculation errors or logical loops on simpler tasks.
These findings connect circuit-level dynamics to macro-level performance, identifying an inherent tension where complex reasoning comes at the cost of elementary computations.
More broadly, our work points to future directions for training policy design, emphasizing the need to balance the development of effective reasoning strategies with the assurance of reliable, flawless execution.
\end{abstract}

\vspace{-10pt}
\section{Introduction}
\vspace{-5pt}
The advent of large reasoning models (LRMs), such as OpenAI o-series~\citep{o1, o3} and DeepSeek-R1~\citep{r1}, has marked a significant milestone in artificial intelligence, demonstrating unprecedented ability in solving complex, multi-step problems.
These models typically employ Chain-of-Thought (CoT) process~\citep{cot}, generating an explicit sequence of reasoning steps before arriving at a final answer.
This capability is substantially enhanced by extensive post-training methods, primarily supervised fine-tuning (SFT) and reinforcement learning (RL)~\citep{trung-etal-2024-reft, xi2024training, mukherjee2025reinforcement}, and by allocating more test-time compute during inference~\citep{zhang2025survey, wu2025inference, snell2025scaling}.

Despite their empirical success, the mechanisms by which these methods enhance reasoning remain largely unclear.
This opacity presents a significant challenge.
For instance, post-trained models often suffer from the ``overthinking problem"~\citep{chen2024not, sui2025stop}, generating excessively long and computationally expensive reasoning chains even for simple tasks, which highlights a critical need for more efficient and adaptive strategies~\citep{tu2025learning, zhang2025continue}.
Furthermore, the community lacks a clear understanding of the fundamental differences between post-training paradigms.
Recent studies have debated whether these methods instill genuinely new problem-solving skills or merely amplify latent capabilities already present in the base model~\citep{rajani2025scalpel, yue2025does, ma2025reasoning}.
Motivated by these trade-offs, several works have proposed ``Think On/Off" controls to manually modulate reasoning depth~\citep{wu2025effectively, qwen3, gpt5}.
However, without a granular understanding of how post-training alters a model's internal mechanism, efforts to improve reasoning are confined to trial-and-error adjustments of training data and resources~\citep{mukherjee2025reinforcement}.

\begin{wrapfigure}{r}{0.47\textwidth}
  \begin{center}
  \vspace{-20pt}%
    \includegraphics[width=0.47\textwidth]{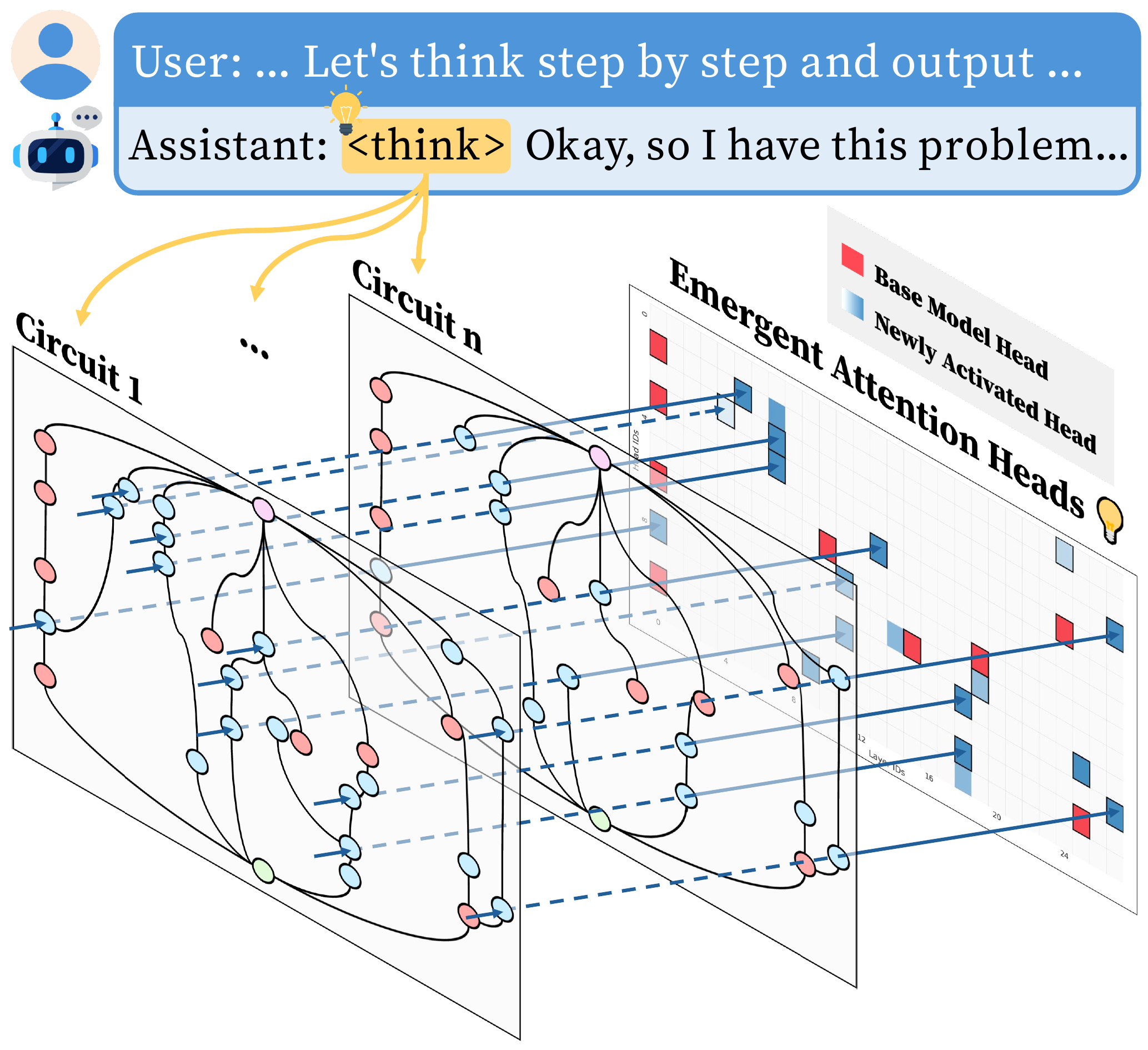}
  \end{center}
  \vspace{-8pt}%
  \caption{\small Reasoning circuits trace the internal computations of LRMs at each checkpoint. After post-training, newly activated attention heads influence the performance at those checkpoints.}
  \vspace{-20pt}
  \label{fig:overview}
\end{wrapfigure}

In this work, we bridge this gap by shifting the analysis from high-level performance metrics to a low-level mechanistic investigation of the model's internal workings.
We employ circuit analysis, a powerful tool of mechanistic interpretability, to identify and characterize functional subgraphs within the transformer architecture~\citep{vaswani2017attention} that are responsible for specific behaviors~\citep{zoom, elhage2021mathematical, wang2023interpretability, bereska2024mechanistic, lindsey2025biology}.
As shown in Figure~\ref{fig:overview}, by applying these lens, we trace the formation of specialized groups of attention heads through reasoning circuits that emerge as a direct consequence of post-training procedures.
This direction is motivated by preliminary findings that particular attention heads correlate with the quality and length of a model's reasoning~\citep{voita-etal-2019-analyzing, cabannes2024iteration, reddy2024the}.

Our investigation and ablation study yield clear, mechanistically insightful findings:
\begin{enumerate}
    \item Distillation and SFT: 
    We find that distillation and SFT induce a large amount of newly emergent heads in circuits.
    Distillation heads are mostly found in early-mid layers, whereas SFT heads are focused on mid-to-late layers.
    They effectively instill complex reasoning with a considerable proportion of attention heads, which also have a potential of confusion.
    \item Group Relative Policy Optimization (GRPO): A prominent RL algorithm, GRPO, engages in dynamic search for reasoning attention heads during the training process, mirroring fluctuations of the task reward signal.
    Its targeted, minimal, but high-impact edits optimize the use of existing knowledge and computational pathways, not building entirely new ones.
    \item Thinking On/Off: While think on mode does not have its own exclusive reasoning heads, think off mode activates enormous attention heads to compensate performance gaps.
    Disabling or scaling down those thinking off heads temporally boosts its performance, but those heads are crucial asset for robust problem solving when the sampling coverage increases.
\end{enumerate}
\vspace{-10pt}

\section{Preliminary}
\label{subsec:circuit-analysis}
\vspace{-5pt}
Transformer circuit models the internal computation of its architecture as a directed acyclic graph (DAG) $G = (\mathcal{N}, E)$, where $\mathcal{N}$ is the set of circuit nodes and a generic node is denoted by $n \in \mathcal{N}$.
Each node corresponds to a distinct component in the model: attention heads $A_{l,j}$ (at layer $l$ and head $j$), MLP modules $M_l$ for each layer, the input node $I$ (embeddings), and the output node $O$ (logits), following~\citep{circuitgrokking, conmy2023towards, ameisen2025circuit}:
\begin{equation}
    \mathcal{N} = \{ I, A_{l,j}, M_l, O \}.
\end{equation}
Edges $E \subseteq \mathcal{N} \times \mathcal{N}$ encode how each node's output contributes to later layers' residual stream inputs:
\begin{equation}
    E = \{ (n_x, n_y) \mid n_x, n_y \in \mathcal{N} \}.
\end{equation}

A circuit is defined as a subgraph $C \subseteq (N, E)$ selected to explain a specific behavior, e.g, how certain tokens influence the model's output or how knowledge is stored and elicited~\citep{kc, ou-etal-2025-llms, temporalhead}.
We specifically implement edge attribution patching with integrated gradients (EAP-IG) which improves faithfulness, wherein ablating all non-circuit edges preserve task performance~\citep{eap, eapig}.

Let $(n_u\!\to\!n_v)\in E$ and let $z_u$ and $z'_u$ denote the clean and corrupted activations of node $n_u$'s output into the residual stream, respectively.
We define the input difference along this edge as $\Delta z_u = z_u - z'_u$.
Following the integrated gradients rule, we average gradients along the straight-line path from $z'_u$ to $z_u$.
As the scalar output signal, we apply a task-agnostic divergence $\mathcal{L}(y_{\text{clean}}, y)$ between the model's output logits at the target position under the clean and interpolated activations, typically a $\mathrm{KL}$ divergence.
We then take gradients of this scalar signal with respect to the \emph{input of node $n_v$} (i.e., $n_v$'s pre-activation into the residual stream).
The EAP-IG edge score is
\begin{equation}
\label{eq:eap-ig}
\mathrm{score}(u\!\to\!v)
\;=\;
\Delta z_u
\cdot
\frac{1}{m}\sum_{k=1}^{m}
\left.
\frac{\partial \mathcal{L}\!\big(z' + \tfrac{k}{m}(z - z')\big)}{\partial\, (\text{input of } n_v)}
\right|_{\;z' + \frac{k}{m}(z - z')}\!,
\end{equation}
where \(m\) is the number of Riemann-sum steps approximating the IG path integral.
We rank edges by \eqref{eq:eap-ig} and select a sparse set by \emph{top-\(n\)} selection.
Lastly, we prune isolated nodes and validate faithfulness via post-hoc interventions: ablate all non-circuit edges (e.g., patching to baseline) and check that task performance is preserved.
Detail of scoring is in \S~\ref{subsec:eapig-detail}.

In addition to our EAP-IG-based analysis, we also conduct an analysis using Sparse Feature Circuits~\citep{markssparse}.
Detail of this analysis is presented in \S\ref{subsec:sfc-detail} and Figure~\ref{fig:sparse_feature_circuit}.

\vspace{-5pt}
\section{Identifying Emergent Attention Heads with Circuits}
To systematically compare how different post-training paradigms change a model’s internal mechanisms, we design a rigorous experiment based on circuit analysis.
Our core methodology for identifying reasoning circuits is a practical application of causal analysis, using ablation as a proxy for more complex patching experiments to identify and validate the causal roles of emergent attention head circuits.
Details of the experimental setup are provided in Appendix~\ref{subsec:exp_setup}.

\paragraph{Circuit Mapping.}
For a given task (e.g., solving an AIME problem), we first map the active computational graph for both the baseline model and a post-trained model. 
As the circuit is structured with pairs of prompts, clean and corrupted, we set clean prompts designed to elicit the reasoning behavior by sampling the answer of each model category.
\begin{itemize}
    \item Baseline model: Answers such as ``To determine the molecular ..." or ``We'll use Python to help us solve ..." for clean, while reasoning model's answer become corrupted.
    Samples can be found in Appendix~\ref{subsec:circuit_setup}.
    \item Reasoning model: Answers right after \texttt{<think>} such as ``Okay, so I have this problem ..." and ``Alright, so I need to find ..." for clean, while baseline model's answer become corrupted.
    Samples can be found in the same \S\ref{subsec:circuit_setup}.
\end{itemize}
Figure~\ref{fig:circuit_example} to~\ref{fig:circuit_example_llama} visualize representative circuit examples for specific tasks.
\vspace{-5pt}
\paragraph{Identifying Emergent Components.}
By comparing circuits of the post-trained model to that of the baseline model, we identify the set of ``emergent heads"—those that are active in the post-trained model but not in the baseline. 
These heads represent the structural changes induced by the training process.
Basically, we specifically pick Qwen families for pair comparison.
Table~\ref{table:reasoning-heads} lists these heads.
We also implement our approach on the Llama-3.2-1B-Instruct~\citep{llama3.2}, applying two distinct post-training methods: SFT and GRPO, for more generalizability.
Additionally, further importance based analysis in \S\ref{app:detail_effect} and Figure~\ref{fig:circuit_effect} is qualitatively support our basic emergence based differentiation.

\paragraph{Causal Validation via Ablation.}
To confirm that these emergent heads are causally responsible for the new reasoning capabilities, we perform ablation inference. 
We run the post-trained model on the evaluation benchmarks but surgically disable the emergent heads by zeroing out their outputs. 
A difference in performance on the target task, compared to the intact post-trained model, serves as strong causal evidence that these heads form a critical part of the newly acquired reasoning circuits.
Ablation details are provided in \S\ref{app:zero-vs-mean-ablation}.
\vspace{-5pt}
\paragraph{Head Activation Scaling.}
Furthermore, we scale up or down activations of each reasoning head in baseline model with their attention head index (layer num and head num).
We then find out the difference in performance both quantitatively and qualitatively.

In the following sections, our investigation reveals that different post-trainings do more than simply fine-tuning a model’s parameters—they fundamentally reshape its internal architecture by strengthening specific attention heads.

\vspace{-5pt}
\begin{table}[t]
\vspace{-5pt}
\centering
\caption{Reasoning Head Ablation Inference for DeepSeek-R1-Distill-Qwen-1.5B and 7B.
Every performance is measured with pass@1 score with temperature 0.6 and 32k context length.
Each ablation cases make the value of specific attention heads, around 5 to 10 number of heads from its circuit results, into zero for checking its importance for reasoning tasks.
We color some scores into red which is the most degraded results except no ablation baseline, while the bold is the completely ruined performance.
We also color performance increase with green when its heads are ablated.
}
\small
\setlength{\tabcolsep}{4pt}
\begin{tabular}{llcccc}
\toprule
\textbf{Model} & \textbf{Method} & \textbf{AIME'24} & \textbf{AIME'25} & \textbf{GPQA} & \textbf{AMC} \\
\midrule
 & No Ablation & 30.0 & 26.6 & 18.6 & 66.2 \\
\multirow{1}{*}{DeepSeekR1-Distill} & Ablation with Reasoning Heads& \textcolor{red}{26.6} & \textcolor{red}{16.6} & 17.1 & 59.0 \\
\multirow{1}{*}{Qwen-1.5B} & Ablation with Base Model Heads & 30.0 & 23.3 & \textcolor{red}{12.1} & \textcolor{red}{53.0} \\
 & Ablation with TriviaQA Heads & \badmetric{0.00} & \badmetric{0.00} & \badmetric{0.00} & \badmetric{0.00} \\
\midrule
 & No Ablation & 40.0 & 43.3 & 35.3 & 81.9 \\
\multirow{1}{*}{DeepSeekR1-Distill} & Ablation with Reasoning Heads & \goodmetric{53.3} & \goodmetric{46.6} & 35.8 & \textcolor{red}{78.3} \\
\multirow{1}{*}{Qwen-7B} & Ablation with Base Model Heads & \goodmetric{53.3} & 43.3 & \goodmetric{37.3} & \goodmetric{83.1} \\
 & Ablation with TriviaQA Heads & 50.0 & 50.0 & \textcolor{red}{34.3} & 79.5 \\
\bottomrule
\label{table:abl_deepseek}
\end{tabular}
\vspace{-20pt}
\end{table}
\vspace{-5pt}
\section{In-Depth Analysis on SFT \& Distillation}
\vspace{-5pt}

\subsection{Distillation heads strongly affect performance}
\label{subsec:distill}

Our primary finding is that distillation induces a set of newly emerged and consistently activated attention heads absent from baseline circuits on the same tasks (AIME'24 and AMC) as shown in Table~\ref{table:reasoning-heads}, Figure~\ref{fig:map_of_reasoning}.
Although about two-thirds of baseline attention-head nodes and all baseline MLP nodes remain active after distillation, the number of newly activated heads is still substantial.
These heads extend, rather than replace, existing model machinery, suggesting that distillation adds specialized components on top of the pretrained foundation.
This interpretation is supported by complementary sparse-feature circuit analysis in \S\ref{subsec:sfc-detail}, which shows increased importance of Layer-0 attention and stronger mid-to-late residual/MLP computations after distillation (Figure~\ref{fig:sparse_feature_circuit}).

To test the functional role of these heads, we perform attention-head ablations by deactivating selected emergent reasoning heads in the distilled model and measuring benchmark performance.
As shown in Table~\ref{table:abl_deepseek}, this ablation consistently degrades performance (e.g., AIME'24 pass@1 drops from 30.0 to 26.6).
Although declines are smaller on GPQA and AMC, where fewer emergent heads are identified, the degradation remains meaningful.
We also compare these effects with ablations of other head groups, including base-model-exclusive heads and heads from TriviaQA circuits.
In the 1.5B model, TriviaQA-head ablation is highly disruptive and can collapse scores to zero, whereas in the 7B model, ablating base-model heads can increase overall benchmark performance.
This illustrates that not all attention heads emerging from post-training are important for reasoning, or they can confuse the model when finding the suitable solution.

The results of Table~\ref{table:abl_qwen_base} further strengthen our insights, as Qwen2.5-Math is more sensitive to ablating its base-model heads than the reasoning heads, reversing the trend seen in DeepSeekR1-Distill in Table~\ref{table:abl_deepseek}.
This cross-model asymmetry confirms that the heads identified by our circuits are specific functional units, rather than a single universal pool of attention heads shared across models.

\begin{figure}[t]
\begin{center}
\includegraphics[width=\columnwidth]{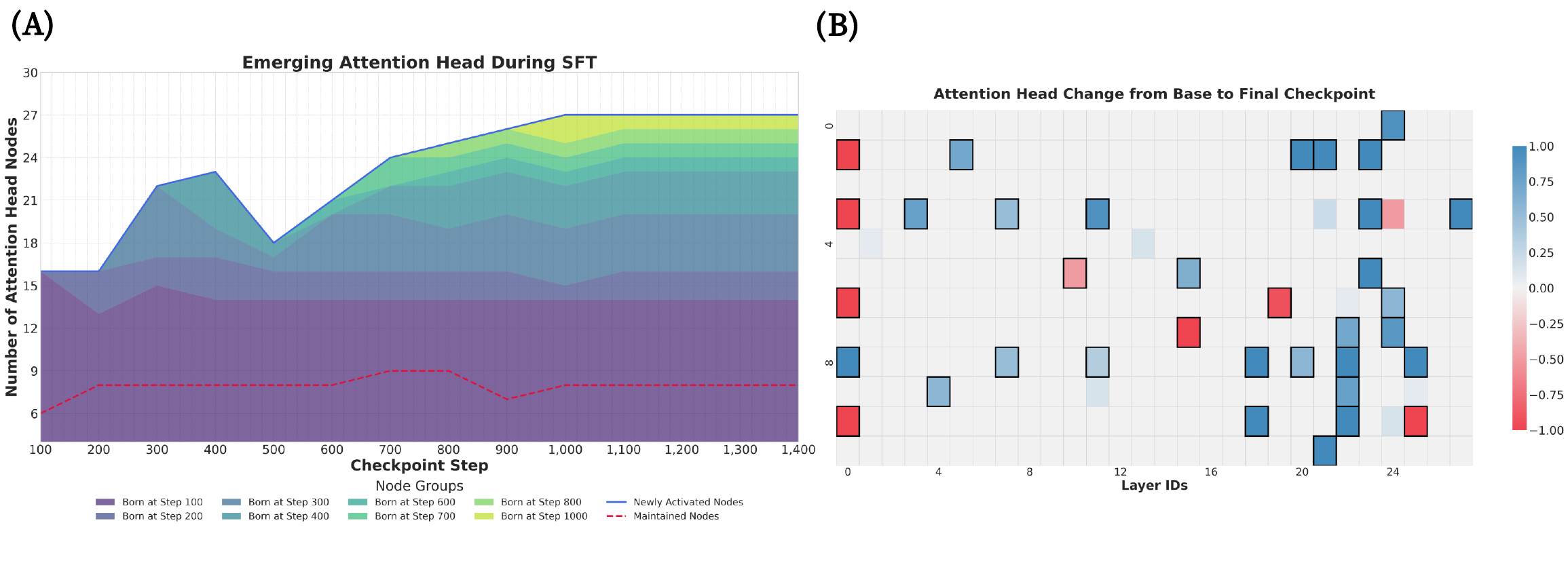}
\end{center}%
\vspace{-10pt}
\caption{Emergent attention heads in Qwen2.5-Math-1.5B during SFT on OpenR1-Math-220k~\citep{openr1}, with circuits constructed on~\cite{aime}.
(A) Cohort analysis across checkpoints. 
The blue curve shows newly activated heads relative to the base model, and the red dashed curve shows retained base-model heads. 
Stacked areas group heads by first-emergence checkpoint, illustrating persistence over training.
(B) Activation-frequency heatmap. 
Red denotes base-model heads (fading with deactivation), and blue denotes emergent heads (darker means higher activation frequency). 
Heads active at the final checkpoint are outlined in black.}
\label{fig:main_fig_sft}
\vspace{-10pt}
\end{figure}

\vspace{-5pt}
\subsection{SFT introduces a large number of attention heads in middle-to-late layers}
\label{subsec:sft}
\vspace{-5pt}
We apply SFT to reasoning traces to approximate a distillation-like training effect.
Following \S\ref{subsec:exp_setup}, we train the baseline model on OpenR1-Math-220k and construct circuits every 100 training steps.
Figures~\ref{fig:main_fig_sft} and~\ref{fig:main_fig_sft_amc} show a pattern similar to DeepSeek distillation.
SFT consistently activates many additional heads, most of which persist to the final checkpoint.
About half emerge by step 100, and most are concentrated in middle-to-late layers.
This persistence pattern suggests that SFT steadily builds new reasoning pathways rather than transient activations.

\paragraph{Quantitative Analysis with Qwen series.}
Using the same ablation protocol as in \S\ref{subsec:distill}, we ablate these SFT-emergent mid-to-late-layer heads.
Ablating about 10 such heads drives every benchmark performances sharply downward, often close to zero.
This trend appears consistently across checkpoints, regardless of baseline checkpoint quality.
We also scale the activations of these heads in the baseline model by a factor of 1.3 to test their functional effect.
The result is a clear trade-off: MATH improves, AMC drops slightly, and AIME'24 still degrades substantially.
Detailed results are reported in Table~\ref{table:abl_sft}.

\paragraph{Qualitative Analysis with Qwen series.}
Comparing newly solved and newly missed items at each checkpoint clarifies this trade-off.
After SFT, models often overcomplicate solutions, replacing short algebraic manipulations with long substitutions or theory-first detours.
As a result, newly introduced errors outnumber newly resolved cases, producing net degradation.
These patterns suggest that SFT installs a more procedural reasoning style but harms strategy selection and path efficiency, causing failures on previously solved items.
Examples are provided in Appendix~\ref{app:qualitative_sft} and Appendix~\ref{app:qualitative_sft_incorrect}.

\paragraph{Re-Implementation with Llama models.}
We repeat the same procedure on Llama-3.2, with results shown in Figure~\ref{fig:main_fig_sft_llama}.
The overall trend matches Qwen: many emergent heads appear and accumulate across checkpoints.
However, their layer distribution differs; unlike Qwen, they are not concentrated in middle-to-late layers.
Instead, they spread across early-to-mid layers, which we hypothesize reflects differences in baseline mathematical capability: Qwen2.5-Math already exhibits nontrivial math competence, whereas Llama-3.2 starts from a weaker arithmetic baseline.

\begin{figure}[t]
\vspace{-10pt}
\begin{center}
\includegraphics[width=\columnwidth]{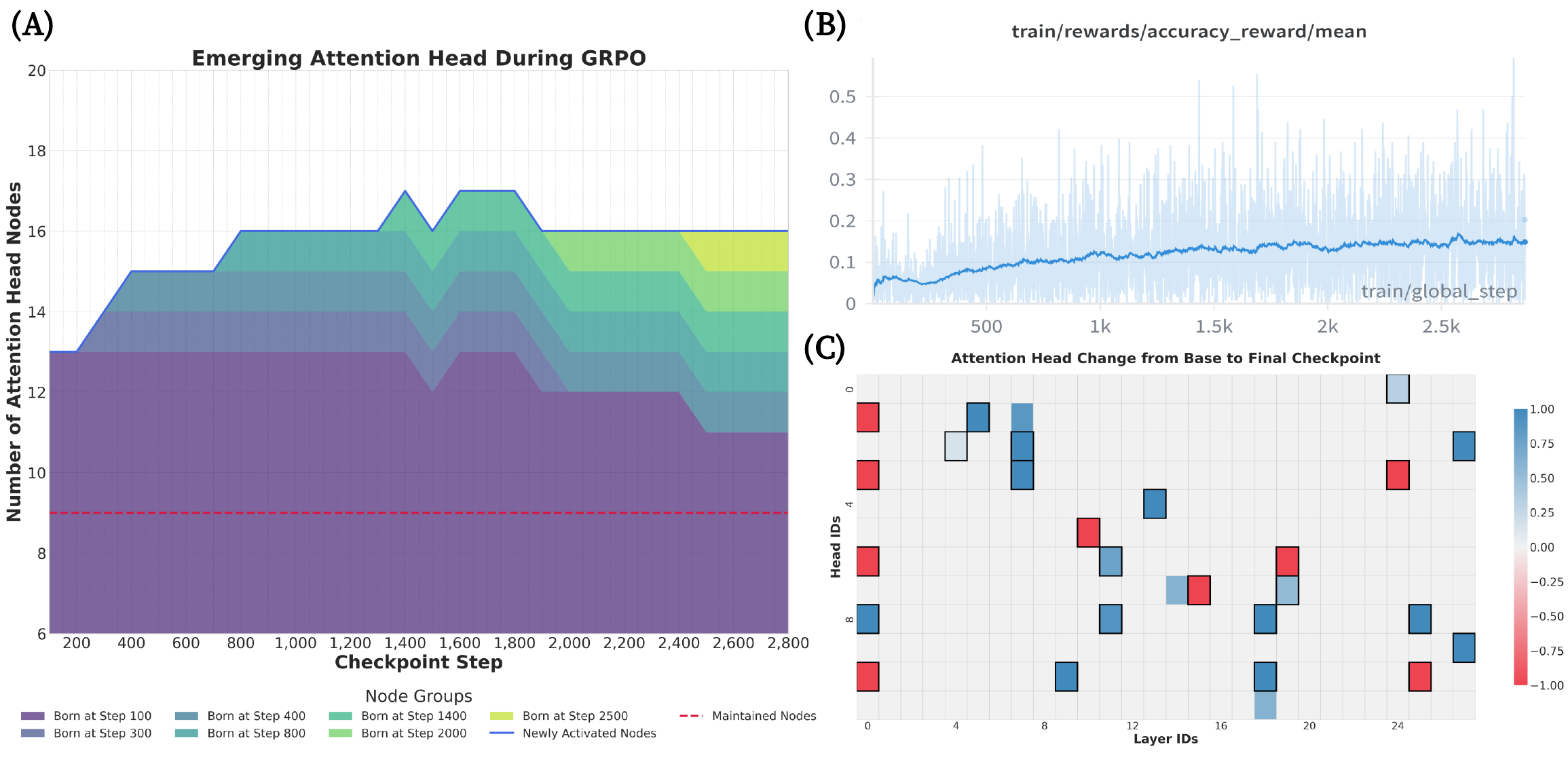}
\end{center}%
\vspace{-10pt}
\caption{Emergent attention heads in Qwen2.5-Math-1.5B during GRPO on OpenR1-Math-220k~\citep{openr1}, with circuits constructed on~\cite{aime}.
The figure follows the same visualization protocol as Figure~\ref{fig:main_fig_sft}.
(A) Cohort analysis across checkpoints; the number of newly activated heads fluctuates with the accuracy-reward trend in (B).
(C) Activation-frequency heatmap with the same color convention as Figure~\ref{fig:main_fig_sft} (red for base-model, blue for emergent heads); heads active at the final checkpoint are outlined in black.
}
\label{fig:main_fig_grpo_math220}
\vspace{-10pt}
\end{figure}
\section{In-Depth Analysis on GRPO}
\label{sec:grpo_analysis}
\paragraph{GRPO helps to find the optimal reasoning path.}
Unlike the relatively static emergent-head pattern observed after SFT, GRPO exhibits a dynamic and reward-driven process of architectural refinement.
Emergent heads are not fixed; they are continuously activated, retained, or pruned according to their reward contribution.
As in the SFT setting, we train the baseline model on OpenR1-Math-220k and additionally on GSM8K (\S\ref{subsec:exp_setup}).
We construct circuits every 100 training steps for each dataset; AIME'24 results are shown in Figure~\ref{fig:main_fig_grpo_math220} and Figure~\ref{fig:main_fig_grpo_gsm8k}.
Learning-rate variants are reported in Figure~\ref{fig:main_fig_grpo_math220_lr2}, and AMC circuits in Figure~\ref{fig:main_fig_grpo_math220_amc}.
Temporal analysis shows that the active-head set remains in constant flux.
In Figure~\ref{fig:main_fig_grpo_math220} (A), the number of newly activated heads oscillates throughout training and tracks the reward-accuracy curve in Figure~\ref{fig:main_fig_grpo_math220} (B).
Heads that emerge early are later pruned when they fail to sustain positive reward impact, while new heads continue to be explored.
This pattern indicates an iterative search for an effective circuit configuration.
The final GRPO head set is small and targeted, with limited overlap with SFT mid-to-late heads, suggesting distinct functional specialization across the two training paradigms.

These dynamics directly reflect the explore--exploit trade-off in reinforcement learning.
Activating a new head can be interpreted as an exploratory test of a candidate computational strategy.
Retaining or pruning that head based on reward impact is exploitation, through which the model refines its functional architecture.

This circuit-level view helps explain why RL acts as a scalpel~\citep{rajani2025scalpel}, producing sparse head updates.
Rather than overwriting the model wholesale, GRPO performs a targeted search for minimal but high-impact functional edits~\citep{mukherjee2025reinforcement}.
It also helps explain why RL-trained capability often remains bounded by the base model's potential~\citep{yue2025does}: GRPO primarily re-optimizes existing knowledge and pathways instead of building new ones from scratch.

\paragraph{Quantitative Analysis with Qwen series.}
Following the intervention setup in Section~\ref{subsec:sft}, we scale GRPO-emergent reasoning heads to quantify their task-specific effects.
Increasing head activation by 1.3\(\times\) for heads from the 100-step GSM8K circuit improves MATH (56\(\rightarrow\)60) but reduces AIME'24 and AMC.
For the single head from the 2500-step OpenR1-Math-220k circuit, 1.3\(\times\) scaling slightly decreases AMC, slightly improves MATH, and leaves AIME'24 nearly unchanged.
In contrast, halving those 100-step GSM8K heads sharply drops AIME'24 (13.3\(\rightarrow\)3.3) while improving MATH (56\(\rightarrow\)63) and AMC (38.5\(\rightarrow\)42.1).
These results indicate strong task-specific trade-offs: heads that benefit one benchmark can impair others.
Because many heads in the 100-step GSM8K circuit seem aligned with AIME-style behavior, suppressing them hurts AIME most but can improve performance on other benchmarks.
Detailed scores are reported in Table~\ref{table:abl_sft}.

\begin{figure}[t]
\vspace{-10pt}
\begin{center}
\includegraphics[width=\columnwidth]{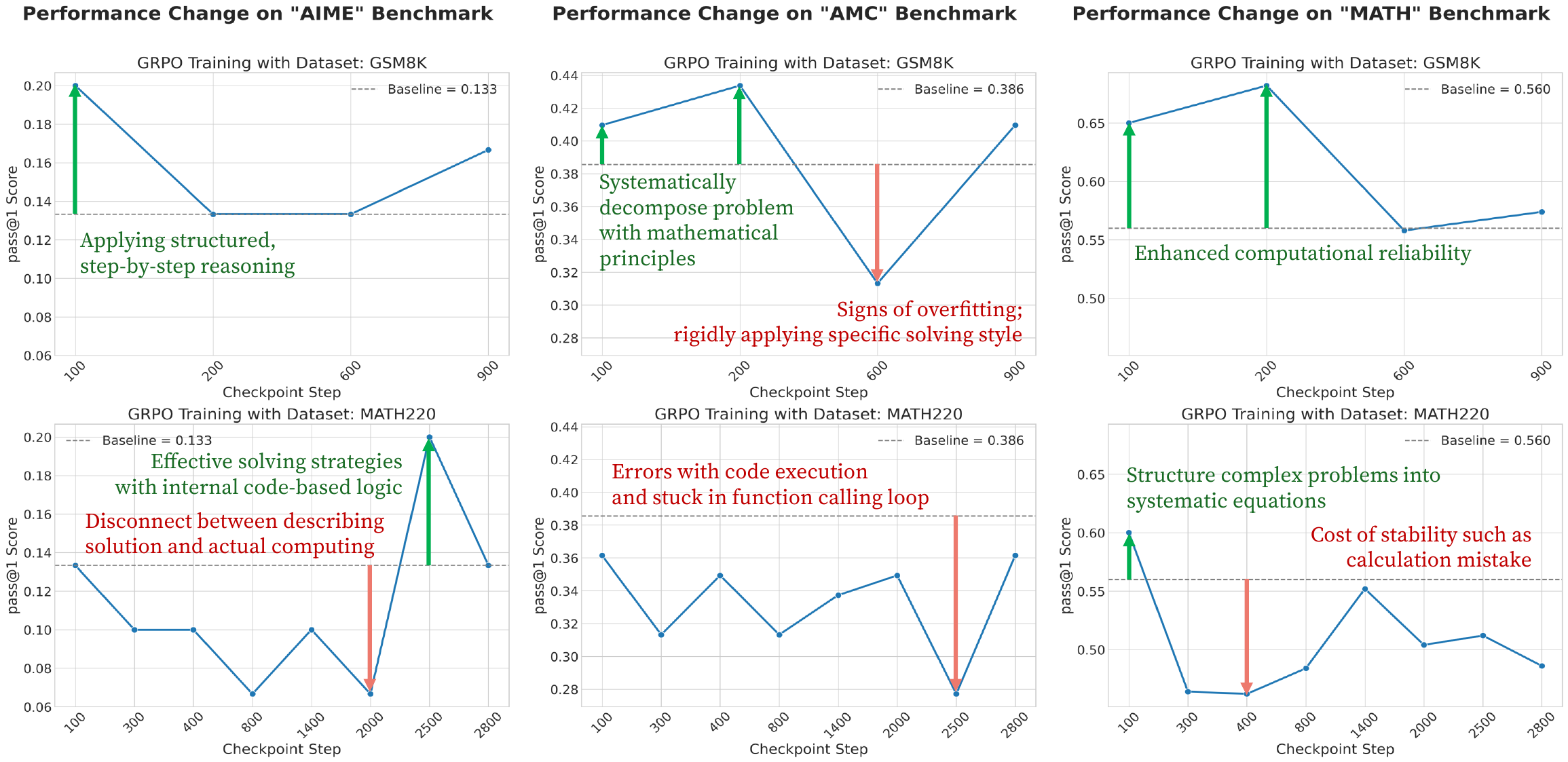}
\end{center}%
\vspace{-10pt}
\caption{Performance change among various benchmarks for each checkpoints of GRPO training with two different training dataset: GSM8K~\citep{gsm8k} and OpenR1-Math-220k~\citep{openr1}.
The green and red arrow indicate impressive performance gain and lose among various checkpoints, and the captions are the summaries of qualitative analysis.
The performance trade-off of each checkpoints is similarly reproduced when we apply attention head scaling with emergent reasoning heads for the baseline model.
Actual examples are presented in the Appendix~\ref{app:qualitative_analysis} to~\ref{app:qualitative_analysis_incorrect}.
}
\label{fig:perform}
\vspace{-10pt}
\end{figure}

Figure~\ref{fig:perform} further illustrates checkpoint-level performance trade-offs after GRPO.
With GSM8K training, performance peaks early (100--200 steps): AIME and AMC increase from 13.3 and 38.6 to about 20 and 43, and MATH rises from 56 to about 67.
Later checkpoints lose part of these gains, consistent with overfitting (e.g., rigid reuse of narrow solution patterns).
With OpenR1-Math-220k training, AIME is highly unstable across checkpoints, alternating between effective strategies and failure modes such as function-calling loops.

\vspace{-5pt}
\paragraph{Qualitative Analysis with Qwen series.}
Compared with the baseline model, GRPO improves multi-step reasoning and problem structuring, especially on composite word problems, often yielding stronger symbolic manipulation and fewer end-stage slips.
However, it can degrade basic numeracy, execution stability, and flexibility in choosing simple programmatic verification when appropriate.
Early checkpoints for both OpenR1-Math-220k and GSM8K improve symbolic manipulation and reduce late-stage errors, but often favor cumbersome analytic derivations over simple programmatic checks.
Mid-to-late checkpoints, which generally score lower, show signs of overfitting and partial forgetting in core algebra and geometry.
Overall, GRPO produces clearer and more systematic reasoning traces and better strategy formation, but may reduce numeracy and robustness when optimization pressure or dataset style dominates.
Representative examples are provided in Appendix~\ref{app:qualitative_grpo} and~\ref{app:qualitative_grpo_incorrect}.
\vspace{-5pt}
\paragraph{Re-Implementation with Llama models.}
We apply the same analysis to Llama-3.2, with results in Figure~\ref{fig:main_fig_grpo_math220_llama}.
The overall trend resembles Qwen2.5-Math: heads emerge and disappear across checkpoints as training searches for effective reasoning pathways.
However, unlike Qwen2.5-Math GRPO, Llama-3.2 GRPO activates many more heads after checkpoint 500, distributed across early-to-mid layers rather than concentrated at a few positions.
We hypothesize that this pattern reflects Llama-3.2's weaker base capability: instead of sharpening a compact reasoning circuit, GRPO appears to allocate capacity to broader skill lifting.
Accordingly, head usage looks closer to SFT-like diffusion than to a compact circuit, though this remains a correlational observation rather than a causal claim.

\begin{table}[t]
\vspace{-5pt}
\centering
\caption{Emergent head ablation inference for Qwen3-8B.
Every performance is measured with pass@1 score with temperature 0.6 and 32k context length, as~\cite{qwen3} suggested for the best performance setting.
Each ablation cases make the value of specific attention heads, around 5 to 10 number of heads from its circuit results, into zero or scale down to half for checking its importance for reasoning tasks.
As no other reasoning heads are found among thinking mode, we do ablation only for thinking off mode.
We color some scores into red for the most degraded results and green for the most performance improvement.
Ablating overstuffed attention heads in thinking off mode increases the baseline score with minimal performance trade-offs.
}
\small
\begin{tabular}{llccccc}
\toprule
\textbf{Model} & \textbf{Method} & \textbf{AIME'24} & \textbf{AIME'25} & \textbf{AMC} & \textbf{GPQA} & \textbf{MATH} \\
\midrule
\multirow{4}{*}{Qwen3-8B} & Think On & 80.0 & 73.3 & 89.1 & 63.1 & 93.8 \\
\cmidrule(lr){2-7}
& Think Off & 30.0 & \textcolor{red}{13.3} & \goodmetric{67.4} & \textcolor{red}{44.9} & \textcolor{red}{81.4} \\
& Think Off \& Ablation & \goodmetric{36.6} & 20.0 & 61.4 & 49.4 & \goodmetric{83.6} \\
& Think Off \& Scale Down & \textcolor{red}{20.0} & \goodmetric{23.3} & \textcolor{red}{56.6} & \goodmetric{51.0} & 81.8 \\
\bottomrule
\label{table:qwen3_abl}
\end{tabular}
\vspace{-10pt}
\end{table}
\section{In-Depth Analysis on Think On/Off}

Recently introduced thinking on/off functionality in models provides a unique window into how efficient reasoning is implemented~\citep{tu2025learning}.
Approaches to reasoning-level control differ across architectures, including system-level routing between fast and deeper models~\citep{gpt5} and system-message keywords that modulate reasoning depth~\citep{oss}.
In this work, we use Qwen3-8B~\citep{qwen3}, whose instruct-style template explicitly gates thinking with the \texttt{<think>} token, enabling clean think-on versus think-off circuit comparisons.

\paragraph{Think-Off Compensation via Broad Head Recruitment.}
Our analysis suggests that think-on triggering in the chat template does not activate a monolithic set of reasoning heads; instead, it selects efficient pathways from the broader attention-head pool.
Circuits extracted from default think-on runs are therefore not unique reasoning-only modules and largely overlap with think-off circuits.
When thinking is disabled via the predefined \verb|<think>\n</think>| template, the model activates many more attention heads.
This pattern indicates that the model has internalized an efficient mechanism for pathway selection.

This behavior differs from post-training settings such as GRPO, where reasoning-specific heads have been observed to emerge.
Because Qwen3 integrates general instruction following (think-off) and reasoning (think-on), it appears to learn a resource-efficient path and to compensate for its removal by recruiting a broader, more redundant head set.
By contrast, think-on engages a more optimized circuit already embedded in the model.

\paragraph{Results of Head Interventions.}
Table~\ref{table:qwen3_abl} reports benchmark performance under different head interventions.
Specifically, we implement attention-head ablation and activation scaling for heads found exclusively in think-off circuits.
Without thinking mode, the model's performance drops substantially, especially on difficult benchmarks such as AIME.
We find that, in think-off mode, ablating a subset of these heads can improve performance across multiple benchmarks, suggesting that removing apparently overactivated or redundant heads may help clarify the model's reasoning pathways.
The largest gains are observed on AIME'24 and AIME'25, which require more complex and structured mathematical reasoning than the other benchmarks.
Meanwhile, scaling down the activation of think-off-specific heads by half also leads to performance gains, and in some cases, such as GPQA and AIME'25, outperforms ablation.
However, this intervention also introduces a trade-off: for example, the AIME'24 score decreases from 30 to 20.

\begin{wrapfigure}{r}{0.5\textwidth}
  \begin{center}
  \vspace{-15pt}%
    \includegraphics[width=0.5\textwidth]{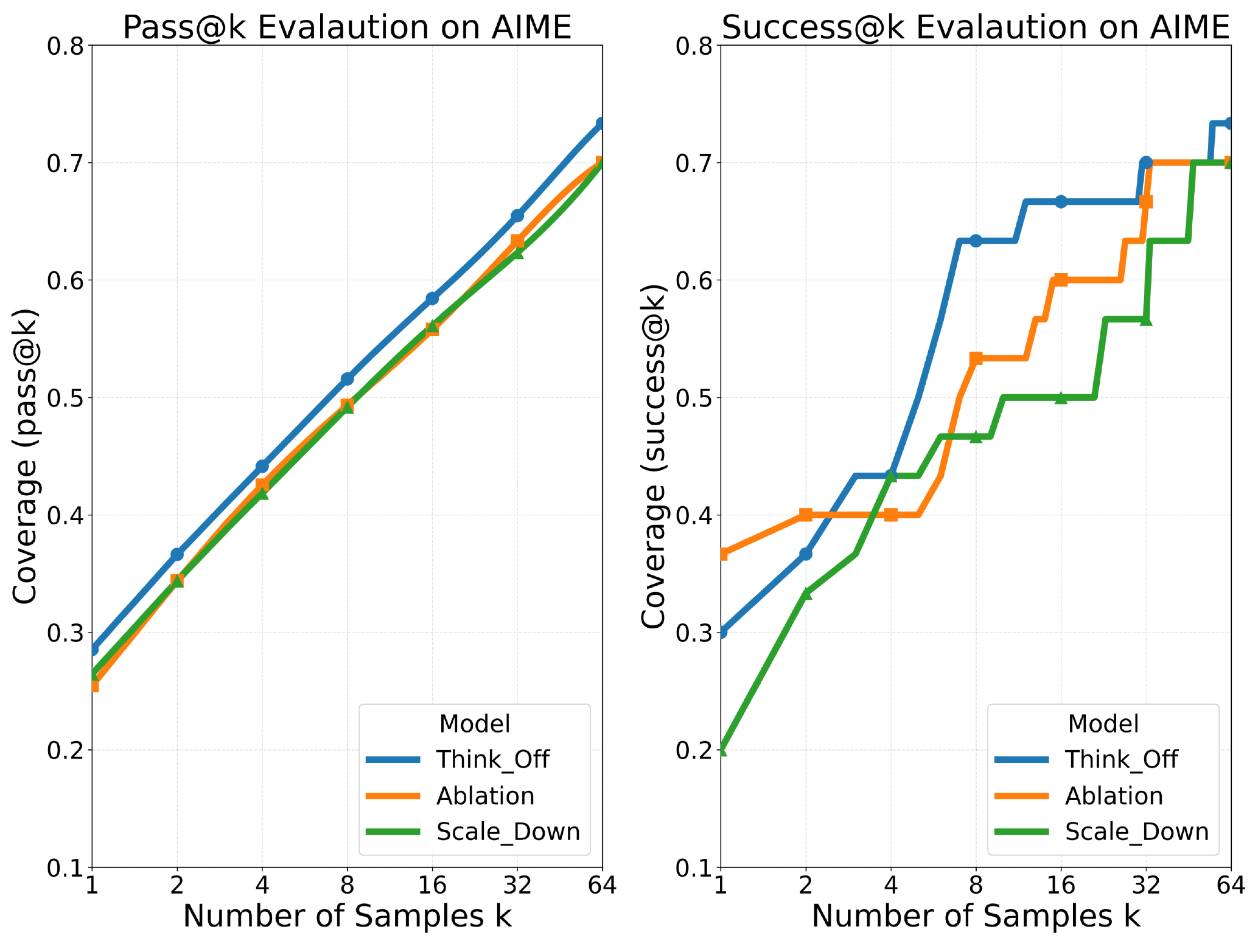}
  \end{center}
  \vspace{-8pt}%
  \caption{Performance differences as sampling coverage increases.
  The left panel shows changes in pass@k with larger coverage, while the right panel reports generation efficiency via success@k.
  }
  \vspace{-10pt}
  \label{fig:passk}
\end{wrapfigure}

\paragraph{Performance Under Increasing Sampling Coverage.}
To further investigate performance under varying sampling coverage, we compare the models' pass@k scores on AIME'24 with up to 64 samples. 
Metric details are provided in \S\ref{app:detail_eval}.
As shown in Figure~\ref{fig:passk} (left), the baseline think-off model consistently maintains a slight performance advantage as $k$ increases. 
We hypothesize that its large number of active attention heads facilitates the exploration of diverse reasoning pathways, a benefit that scales with the number of samples. 
In contrast, the ablated and scaled-down models show reduced ability to discover novel solutions at higher $k$ and larger sample counts $n$. 
This behavior is reminiscent of models that, after post-training like GRPO, become locked into specific reasoning paths and fail to solve certain problems regardless of the increased coverage~\citep{yue2025does}.

This trade-off is clearer in generation efficiency (success@k), which measures the probability of finding a correct solution within each trial (Figure~\ref{fig:passk}, right).
Here, the ablation model initially outperforms the baseline at very low sampling rates ($k \le 2$), suggesting that simplification of attention heads helps focus the model on a more direct and efficient reasoning path. 
However, this advantage quickly disappears as $k$ increases, where the baseline benefits more from broader solution-space exploration.
Meanwhile, the scaled-down model consistently underperforms, lacking both the focused efficiency of ablation and the exploratory breadth of the baseline.
Collectively, these results highlight the dual role of the numerous emergent heads in think-off mode: they can add noise in low-sample settings but become valuable for robust problem-solving under larger sampling.
\vspace{-10pt}
\section{Conclusion, Limitation, and Future Work}
\vspace{-10pt}
We present comparative, mechanistic account of how post-training paradigms reconfigure the internal mechanism of reasoning models.
Our analyses show that these methods do not merely explore a fixed parameter landscape, instead, they reshape functional structure:
distillation and SFT steadily embed new computational pathways via the sustained emergence of additional, large reasoning heads, on the other hand, GRPO conducts reward-guided head configurations, with heads appearing and being pruned over training, to optimize capabilities. 
The think on/off architecture behaves as a selective gate, as thinking mode activates just the task-relevant heads, while thinking off compensates ability through more diverse attentions with enormous heads.
And their differences align with observed performance trade-offs: 
the systems more often solve hard problems by forming deeper, more structured plans, yet sometimes regress on previously easy items due to over reasoning or arithmetic slips. 

Although this provides a new lens through which to view post-training,
our findings are constrained by two factors. 
First, the generalizability of our implementation has only been validated on the Qwen model series and single Llama model.
Although our re-implementation on Llama confirms a relatively effective transition, further work is necessary to establish its effectiveness across a broader spectrum of model architectures.
Second, our analysis relies on prompt-based circuits, which demand precise setup and may be vulnerable to polysemanticity. 
While alternative approaches like SAE-based circuits could mitigate this issue, we deemed them impractical for this study, as they are computationally costly and less generalizable, requiring separate SAEs to be trained for every checkpoints. 

Still, its conclusions are subject to offer avenues for future research.
Taken together, our results motivate attention head informed training policies that
(i) encourage targeted head activation rather than uncontrolled head growth, 
(ii) use reward shaping to jointly optimize plan quality and calculation reliability, 
and (iii) leverage per-head influence estimates to guide selective post-training. 
We view this mechanistic perspective as a foundation for principled, interpretable, and robust post-training of effective reasoning strategies with the assurance of reliable, flawless execution.

\subsection*{Acknowledgments}
We thank Taewhoo Lee for the valuable feedback on our work.
This work was supported in part by the National Research Foundation of Korea [NRF-2023R1A2C3004176, RS-2023-00262002], the Ministry of Health \& Welfare, Republic of Korea [HR20C002103], and the ICT Creative Consilience program through the Institute of Information \& Communications Technology Planning \& Evaluation (IITP) grant funded by the MSIT [IITP-2025-2020-0-01819].

\bibliography{colm2026_conference}

@article{o1,
  title={Openai o1 system card},
  author={Jaech, Aaron and Kalai, Adam and Lerer, Adam and Richardson, Adam and El-Kishky, Ahmed and Low, Aiden and Helyar, Alec and Madry, Aleksander and Beutel, Alex and Carney, Alex and others},
  journal={arXiv preprint arXiv:2412.16720},
  year={2024}
}

@article{o3,
title = {OpenAI o3 and o4-mini System Card},
author = {OpenAI},
year = {2025}
}

@article{gpt5,
title = {GPT-5 System Card},
author = {OpenAI},
year = {2025}
}

@article{r1,
  title={Deepseek-r1: Incentivizing reasoning capability in llms via reinforcement learning},
  author={Guo, Daya and Yang, Dejian and Zhang, Haowei and Song, Junxiao and Zhang, Ruoyu and Xu, Runxin and Zhu, Qihao and Ma, Shirong and Wang, Peiyi and Bi, Xiao and others},
  journal={arXiv preprint arXiv:2501.12948},
  year={2025}
}

@article{cot,
  title={Chain-of-thought prompting elicits reasoning in large language models},
  author={Wei, Jason and Wang, Xuezhi and Schuurmans, Dale and Bosma, Maarten and Xia, Fei and Chi, Ed and Le, Quoc V and Zhou, Denny and others},
  journal={Advances in neural information processing systems},
  volume={35},
  pages={24824--24837},
  year={2022}
}

@article{qwen3,
  title={Qwen3 technical report},
  author={Yang, An and Li, Anfeng and Yang, Baosong and Zhang, Beichen and Hui, Binyuan and Zheng, Bo and Yu, Bowen and Gao, Chang and Huang, Chengen and Lv, Chenxu and others},
  journal={arXiv preprint arXiv:2505.09388},
  year={2025}
}

@inproceedings{
lightman2024lets,
title={Let's Verify Step by Step},
author={Hunter Lightman and Vineet Kosaraju and Yuri Burda and Harrison Edwards and Bowen Baker and Teddy Lee and Jan Leike and John Schulman and Ilya Sutskever and Karl Cobbe},
booktitle={The Twelfth International Conference on Learning Representations},
year={2024},
url={https://openreview.net/forum?id=v8L0pN6EOi}
}

@inproceedings{
wu2025inference,
title={Inference Scaling Laws: An Empirical Analysis of Compute-Optimal Inference for {LLM} Problem-Solving},
author={Yangzhen Wu and Zhiqing Sun and Shanda Li and Sean Welleck and Yiming Yang},
booktitle={The Thirteenth International Conference on Learning Representations},
year={2025},
url={https://openreview.net/forum?id=VNckp7JEHn}
}

@inproceedings{
snell2025scaling,
title={Scaling {LLM} Test-Time Compute Optimally Can be More Effective than Scaling Parameters for Reasoning},
author={Charlie Victor Snell and Jaehoon Lee and Kelvin Xu and Aviral Kumar},
booktitle={The Thirteenth International Conference on Learning Representations},
year={2025},
url={https://openreview.net/forum?id=4FWAwZtd2n}
}

@article{zhang2025survey,
  title={A Survey on Test-Time Scaling in Large Language Models: What, How, Where, and How Well?},
  author={Zhang, Qiyuan and Lyu, Fuyuan and Sun, Zexu and Wang, Lei and Zhang, Weixu and Hua, Wenyue and Wu, Haolun and Guo, Zhihan and Wang, Yufei and Muennighoff, Niklas and others},
  journal={arXiv preprint arXiv:2503.24235},
  year={2025}
}

@article{tu2025learning,
  title={Learning When to Think: Shaping Adaptive Reasoning in R1-Style Models via Multi-Stage RL},
  author={Tu, Songjun and Lin, Jiahao and Zhang, Qichao and Tian, Xiangyu and Li, Linjing and Lan, Xiangyuan and Zhao, Dongbin},
  journal={arXiv preprint arXiv:2505.10832},
  year={2025}
}

@article{zhang2025continue,
  title={When to continue thinking: Adaptive thinking mode switching for efficient reasoning},
  author={Zhang, Xiaoyun and Ruan, Jingqing and Ma, Xing and Zhu, Yawen and Zhao, Haodong and Li, Hao and Chen, Jiansong and Zeng, Ke and Cai, Xunliang},
  journal={arXiv preprint arXiv:2505.15400},
  year={2025}
}

@article{rajani2025scalpel,
  title={Scalpel vs. Hammer: GRPO Amplifies Existing Capabilities, SFT Replaces Them},
  author={Rajani, Neel and Gema, Aryo Pradipta and Goldfarb-Tarrant, Seraphina and Titov, Ivan},
  journal={arXiv preprint arXiv:2507.10616},
  year={2025}
}

@article{mukherjee2025reinforcement,
  title={Reinforcement Learning Finetunes Small Subnetworks in Large Language Models},
  author={Mukherjee, Sagnik and Yuan, Lifan and Hakkani-Tur, Dilek and Peng, Hao},
  journal={arXiv preprint arXiv:2505.11711},
  year={2025}
}

@article{yue2025does,
  title={Does reinforcement learning really incentivize reasoning capacity in llms beyond the base model?},
  author={Yue, Yang and Chen, Zhiqi and Lu, Rui and Zhao, Andrew and Wang, Zhaokai and Song, Shiji and Huang, Gao},
  journal={arXiv preprint arXiv:2504.13837},
  year={2025}
}

@article{wu2025effectively,
  title={Effectively controlling reasoning models through thinking intervention},
  author={Wu, Tong and Xiang, Chong and Wang, Jiachen T and Suh, G Edward and Mittal, Prateek},
  journal={arXiv preprint arXiv:2503.24370},
  year={2025}
}

@inproceedings{xi2024training,
  title={Training Large Language Models for Reasoning through Reverse Curriculum Reinforcement Learning},
  author={Xi, Zhiheng and Chen, Wenxiang and Hong, Boyang and Jin, Senjie and Zheng, Rui and He, Wei and Ding, Yiwen and Liu, Shichun and Guo, Xin and Wang, Junzhe and others},
  booktitle={International Conference on Machine Learning},
  pages={54030--54048},
  year={2024},
  organization={PMLR}
}

@inproceedings{trung-etal-2024-reft,
    title = "{R}e{FT}: Reasoning with Reinforced Fine-Tuning",
    author = "Trung, Luong  and
      Zhang, Xinbo  and
      Jie, Zhanming  and
      Sun, Peng  and
      Jin, Xiaoran  and
      Li, Hang",
    editor = "Ku, Lun-Wei  and
      Martins, Andre  and
      Srikumar, Vivek",
    booktitle = "Proceedings of the 62nd Annual Meeting of the Association for Computational Linguistics (Volume 1: Long Papers)",
    month = aug,
    year = "2024",
    address = "Bangkok, Thailand",
    publisher = "Association for Computational Linguistics",
    url = "https://aclanthology.org/2024.acl-long.410/",
    doi = "10.18653/v1/2024.acl-long.410",
    pages = "7601--7614",
}

@inproceedings{
wang2023interpretability,
title={Interpretability in the Wild: a Circuit for Indirect Object Identification in {GPT}-2 Small},
author={Kevin Ro Wang and Alexandre Variengien and Arthur Conmy and Buck Shlegeris and Jacob Steinhardt},
booktitle={The Eleventh International Conference on Learning Representations },
year={2023},
url={https://openreview.net/forum?id=NpsVSN6o4ul}
}

@article{elhage2021mathematical,
   title={A Mathematical Framework for Transformer Circuits},
   author={Elhage, Nelson and Nanda, Neel and Olsson, Catherine and Henighan, Tom and Joseph, Nicholas and Mann, Ben and Askell, Amanda and Bai, Yuntao and Chen, Anna and Conerly, Tom and DasSarma, Nova and Drain, Dawn and Ganguli, Deep and Hatfield-Dodds, Zac and Hernandez, Danny and Jones, Andy and Kernion, Jackson and Lovitt, Liane and Ndousse, Kamal and Amodei, Dario and Brown, Tom and Clark, Jack and Kaplan, Jared and McCandlish, Sam and Olah, Chris},
   year={2021},
   journal={Transformer Circuits Thread},
   note={https://transformer-circuits.pub/2021/framework/index.html}
}

@article{
bereska2024mechanistic,
title={Mechanistic Interpretability for {AI} Safety - A Review},
author={Leonard Bereska and Stratis Gavves},
journal={Transactions on Machine Learning Research},
issn={2835-8856},
year={2024},
url={https://openreview.net/forum?id=ePUVetPKu6},
note={Survey Certification, Expert Certification}
}

@article{lindsey2025biology,
  author={Lindsey, Jack and Gurnee, Wes and Ameisen, Emmanuel and Chen, Brian and Pearce, Adam and Turner, Nicholas L. and Citro, Craig and Abrahams, David and Carter, Shan and Hosmer, Basil and Marcus, Jonathan and Sklar, Michael and Templeton, Adly and Bricken, Trenton and McDougall, Callum and Cunningham, Hoagy and Henighan, Thomas and Jermyn, Adam and Jones, Andy and Persic, Andrew and Qi, Zhenyi and Thompson, T. Ben and Zimmerman, Sam and Rivoire, Kelley and Conerly, Thomas and Olah, Chris and Batson, Joshua},
  title={On the Biology of a Large Language Model},
  journal={Transformer Circuits Thread},
  year={2025},
  url={https://transformer-circuits.pub/2025/attribution-graphs/biology.html}
}

@inproceedings{voita-etal-2019-analyzing,
    title = "Analyzing Multi-Head Self-Attention: Specialized Heads Do the Heavy Lifting, the Rest Can Be Pruned",
    author = "Voita, Elena  and
      Talbot, David  and
      Moiseev, Fedor  and
      Sennrich, Rico  and
      Titov, Ivan",
    editor = "Korhonen, Anna  and
      Traum, David  and
      M{\`a}rquez, Llu{\'i}s",
    booktitle = "Proceedings of the 57th Annual Meeting of the Association for Computational Linguistics",
    month = jul,
    year = "2019",
    address = "Florence, Italy",
    publisher = "Association for Computational Linguistics",
    url = "https://aclanthology.org/P19-1580/",
    doi = "10.18653/v1/P19-1580",
    pages = "5797--5808",
}

@article{cabannes2024iteration,
  title={Iteration head: A mechanistic study of chain-of-thought},
  author={Cabannes, Vivien and Arnal, Charles and Bouaziz, Wassim and Yang, Xingyu and Charton, Francois and Kempe, Julia},
  journal={Advances in Neural Information Processing Systems},
  volume={37},
  pages={109101--109122},
  year={2024}
}

@inproceedings{
reddy2024the,
title={The mechanistic basis of data dependence and abrupt learning in an in-context classification task},
author={Gautam Reddy},
booktitle={The Twelfth International Conference on Learning Representations},
year={2024},
url={https://openreview.net/forum?id=aN4Jf6Cx69}
}

@inproceedings{
circuitgrokking,
title={Progress measures for grokking via mechanistic interpretability},
author={Neel Nanda and Lawrence Chan and Tom Lieberum and Jess Smith and Jacob Steinhardt},
booktitle={The Eleventh International Conference on Learning Representations },
year={2023},
url={https://openreview.net/forum?id=9XFSbDPmdW}
}

@article{conmy2023towards,
  title={Towards automated circuit discovery for mechanistic interpretability},
  author={Conmy, Arthur and Mavor-Parker, Augustine and Lynch, Aengus and Heimersheim, Stefan and Garriga-Alonso, Adri{\`a}},
  journal={Advances in Neural Information Processing Systems},
  volume={36},
  pages={16318--16352},
  year={2023}
}

@inproceedings{
eapig,
title={Have Faith in Faithfulness: Going Beyond Circuit Overlap When Finding Model Mechanisms},
author={Michael Hanna and Sandro Pezzelle and Yonatan Belinkov},
booktitle={First Conference on Language Modeling},
year={2024},
url={https://openreview.net/forum?id=TZ0CCGDcuT}
}

@inproceedings{ou-etal-2025-llms,
    title = "How Do {LLM}s Acquire New Knowledge? A Knowledge Circuits Perspective on Continual Pre-Training",
    author = "Ou, Yixin  and
      Yao, Yunzhi  and
      Zhang, Ningyu  and
      Jin, Hui  and
      Sun, Jiacheng  and
      Deng, Shumin  and
      Li, Zhenguo  and
      Chen, Huajun",
    editor = "Che, Wanxiang  and
      Nabende, Joyce  and
      Shutova, Ekaterina  and
      Pilehvar, Mohammad Taher",
    booktitle = "Findings of the Association for Computational Linguistics: ACL 2025",
    month = jul,
    year = "2025",
    address = "Vienna, Austria",
    publisher = "Association for Computational Linguistics",
    url = "https://aclanthology.org/2025.findings-acl.1021/",
    doi = "10.18653/v1/2025.findings-acl.1021",
    pages = "19889--19913",
    ISBN = "979-8-89176-256-5",
}

@inproceedings{temporalhead,
    title = "Does Time Have Its Place? Temporal Heads: Where Language Models Recall Time-specific Information",
    author = "Park, Yein  and
      Yoon, Chanwoong  and
      Park, Jungwoo  and
      Jeong, Minbyul  and
      Kang, Jaewoo",
    editor = "Che, Wanxiang  and
      Nabende, Joyce  and
      Shutova, Ekaterina  and
      Pilehvar, Mohammad Taher",
    booktitle = "Proceedings of the 63rd Annual Meeting of the Association for Computational Linguistics (Volume 1: Long Papers)",
    month = jul,
    year = "2025",
    address = "Vienna, Austria",
    publisher = "Association for Computational Linguistics",
    url = "https://aclanthology.org/2025.acl-long.812/",
    doi = "10.18653/v1/2025.acl-long.812",
    pages = "16616--16643",
    ISBN = "979-8-89176-251-0",
}

@article{ameisen2025circuit,
  author={Ameisen, Emmanuel and Lindsey, Jack and Pearce, Adam and Gurnee, Wes and Turner, Nicholas L. and Chen, Brian and Citro, Craig and Abrahams, David and Carter, Shan and Hosmer, Basil and Marcus, Jonathan and Sklar, Michael and Templeton, Adly and Bricken, Trenton and McDougall, Callum and Cunningham, Hoagy and Henighan, Thomas and Jermyn, Adam and Jones, Andy and Persic, Andrew and Qi, Zhenyi and Ben Thompson, T. and Zimmerman, Sam and Rivoire, Kelley and Conerly, Thomas and Olah, Chris and Batson, Joshua},
  title={Circuit Tracing: Revealing Computational Graphs in Language Models},
  journal={Transformer Circuits Thread},
  year={2025},
  url={https://transformer-circuits.pub/2025/attribution-graphs/methods.html}
}

@article{eap,
	title = {Attribution {Patching}: {Activation} {Patching} {At} {Industrial} {Scale}},
	url = {https://www.neelnanda.io/mechanistic-interpretability/attribution-patching},
	author = {Neel Nanda},
	year={2023},
}

@inproceedings{
wei2022finetuned,
title={Finetuned Language Models are Zero-Shot Learners},
author={Jason Wei and Maarten Bosma and Vincent Zhao and Kelvin Guu and Adams Wei Yu and Brian Lester and Nan Du and Andrew M. Dai and Quoc V Le},
booktitle={International Conference on Learning Representations},
year={2022},
url={https://openreview.net/forum?id=gEZrGCozdqR}
}

@article{qwen2.5math,
  title={Qwen2. 5-math technical report: Toward mathematical expert model via self-improvement},
  author={Yang, An and Zhang, Beichen and Hui, Binyuan and Gao, Bofei and Yu, Bowen and Li, Chengpeng and Liu, Dayiheng and Tu, Jianhong and Zhou, Jingren and Lin, Junyang and others},
  journal={arXiv preprint arXiv:2409.12122},
  year={2024}
}

@article{toshniwal2024openmathinstruct,
  title={Openmathinstruct-1: A 1.8 million math instruction tuning dataset},
  author={Toshniwal, Shubham and Moshkov, Ivan and Narenthiran, Sean and Gitman, Daria and Jia, Fei and Gitman, Igor},
  journal={Advances in Neural Information Processing Systems},
  volume={37},
  pages={34737--34774},
  year={2024}
}

@inproceedings{kard,
 author = {Kang, Minki and Lee, Seanie and Baek, Jinheon and Kawaguchi, Kenji and Hwang, Sung Ju},
 booktitle = {Advances in Neural Information Processing Systems},
 editor = {A. Oh and T. Naumann and A. Globerson and K. Saenko and M. Hardt and S. Levine},
 pages = {48573--48602},
 publisher = {Curran Associates, Inc.},
 title = {Knowledge-Augmented Reasoning Distillation for Small Language Models in Knowledge-Intensive Tasks},
 url = {https://proceedings.neurips.cc/paper_files/paper/2023/file/97faedc90260eae5c400f92d5831c3d7-Paper-Conference.pdf},
 volume = {36},
 year = {2023}
}

@misc{openr1,
    title = {Open R1: A fully open reproduction of DeepSeek-R1},
    url = {https://github.com/huggingface/open-r1},
    author = {{Hugging Face}},
    month = {January},
    year = {2025}
}

@article{shao2024deepseekmath,
  title={Deepseekmath: Pushing the limits of mathematical reasoning in open language models},
  author={Shao, Zhihong and Wang, Peiyi and Zhu, Qihao and Xu, Runxin and Song, Junxiao and Bi, Xiao and Zhang, Haowei and Zhang, Mingchuan and Li, YK and Wu, Yang and others},
  journal={arXiv preprint arXiv:2402.03300},
  year={2024}
}

@article{ppo,
  title={Proximal policy optimization algorithms},
  author={Schulman, John and Wolski, Filip and Dhariwal, Prafulla and Radford, Alec and Klimov, Oleg},
  journal={arXiv preprint arXiv:1707.06347},
  year={2017}
}

@article{ouyang2022training,
  title={Training language models to follow instructions with human feedback},
  author={Ouyang, Long and Wu, Jeffrey and Jiang, Xu and Almeida, Diogo and Wainwright, Carroll and Mishkin, Pamela and Zhang, Chong and Agarwal, Sandhini and Slama, Katarina and Ray, Alex and others},
  journal={Advances in neural information processing systems},
  volume={35},
  pages={27730--27744},
  year={2022}
}

@article{gsm8k,
  title={Training verifiers to solve math word problems},
  author={Cobbe, Karl and Kosaraju, Vineet and Bavarian, Mohammad and Chen, Mark and Jun, Heewoo and Kaiser, Lukasz and Plappert, Matthias and Tworek, Jerry and Hilton, Jacob and Nakano, Reiichiro and others},
  journal={arXiv preprint arXiv:2110.14168},
  year={2021}
}

@inproceedings{triviaqa,
    title = "{T}rivia{QA}: A Large Scale Distantly Supervised Challenge Dataset for Reading Comprehension",
    author = "Joshi, Mandar  and
      Choi, Eunsol  and
      Weld, Daniel  and
      Zettlemoyer, Luke",
    editor = "Barzilay, Regina  and
      Kan, Min-Yen",
    booktitle = "Proceedings of the 55th Annual Meeting of the Association for Computational Linguistics (Volume 1: Long Papers)",
    month = jul,
    year = "2017",
    address = "Vancouver, Canada",
    publisher = "Association for Computational Linguistics",
    url = "https://aclanthology.org/P17-1147/",
    doi = "10.18653/v1/P17-1147",
    pages = "1601--1611",
}

@inproceedings{
gpqa,
title={{GPQA}: A Graduate-Level Google-Proof Q\&A Benchmark},
author={David Rein and Betty Li Hou and Asa Cooper Stickland and Jackson Petty and Richard Yuanzhe Pang and Julien Dirani and Julian Michael and Samuel R. Bowman},
booktitle={First Conference on Language Modeling},
year={2024},
url={https://openreview.net/forum?id=Ti67584b98}
}

@article{lightr1,
  title={Light-R1: Curriculum SFT, DPO and RL for Long COT from Scratch and Beyond},
  author={Wen, Liang and Cai, Yunke and Xiao, Fenrui and He, Xin and An, Qi and Duan, Zhenyu and Du, Yimin and Liu, Junchen and Tang, Lifu and Lv, Xiaowei and Zou, Haosheng and Deng, Yongchao and Jia, Shousheng and Zhang, Xiangzheng},
  journal={arXiv preprint arXiv:2503.10460},
  year={2025}
}

@misc{aime,
      title={{AIME} Problems and Solutions},
      author={{AIME}},
      year={2025},
      url={https://artofproblemsolving.com/wiki/index.php/AIME_Problems_and_Solutions}
}

@misc{AMC2023,
  title        = {{Amc 2023}},
  author       = {AI-MO},
  year         = 2024,
  url = {https://huggingface.co/datasets/AI-MO/
aimo-validation-amc}
}

@article{oss,
  title={gpt-oss-120b \& gpt-oss-20b Model Card},
  author={Agarwal, Sandhini and Ahmad, Lama and Ai, Jason and Altman, Sam and Applebaum, Andy and Arbus, Edwin and Arora, Rahul K and Bai, Yu and Baker, Bowen and Bao, Haiming and others},
  journal={arXiv preprint arXiv:2508.10925},
  year={2025}
}

@inproceedings{
chu2025sft,
title={{SFT} Memorizes, {RL} Generalizes: A Comparative Study of Foundation Model Post-training},
author={Tianzhe Chu and Yuexiang Zhai and Jihan Yang and Shengbang Tong and Saining Xie and Dale Schuurmans and Quoc V Le and Sergey Levine and Yi Ma},
booktitle={Forty-second International Conference on Machine Learning},
year={2025},
url={https://openreview.net/forum?id=dYur3yabMj}
}

@article{ma2025reasoning,
  title={Reasoning models can be effective without thinking},
  author={Ma, Wenjie and He, Jingxuan and Snell, Charlie and Griggs, Tyler and Min, Sewon and Zaharia, Matei},
  journal={arXiv preprint arXiv:2504.09858},
  year={2025}
}

@article{zoom,
  author = {Olah, Chris and Cammarata, Nick and Schubert, Ludwig and Goh, Gabriel and Petrov, Michael and Carter, Shan},
  title = {Zoom In: An Introduction to Circuits},
  journal = {Distill},
  year = {2020},
  note = {https://distill.pub/2020/circuits/zoom-in},
  doi = {10.23915/distill.00024.001}
}

@article{kc,
  title={Knowledge circuits in pretrained transformers},
  author={Yao, Yunzhi and Zhang, Ningyu and Xi, Zekun and Wang, Mengru and Xu, Ziwen and Deng, Shumin and Chen, Huajun},
  journal={Advances in Neural Information Processing Systems},
  volume={37},
  pages={118571--118602},
  year={2024}
}

@article{pass@1,
  title={Evaluating large language models trained on code},
  author={Chen, Mark and Tworek, Jerry and Jun, Heewoo and Yuan, Qiming and Pinto, Henrique Ponde De Oliveira and Kaplan, Jared and Edwards, Harri and Burda, Yuri and Joseph, Nicholas and Brockman, Greg and others},
  journal={arXiv preprint arXiv:2107.03374},
  year={2021}
}

@article{rl_survey,
  title={A Survey of Reinforcement Learning for Large Reasoning Models},
  author={Zhang, Kaiyan and Zuo, Yuxin and He, Bingxiang and Sun, Youbang and Liu, Runze and Jiang, Che and Fan, Yuchen and Tian, Kai and Jia, Guoli and Li, Pengfei and others},
  journal={arXiv preprint arXiv:2509.08827},
  year={2025}
}

@article{
sui2025stop,
title={Stop Overthinking: A Survey on Efficient Reasoning for Large Language Models},
author={Yang Sui and Yu-Neng Chuang and Guanchu Wang and Jiamu Zhang and Tianyi Zhang and Jiayi Yuan and Hongyi Liu and Andrew Wen and Shaochen Zhong and Na Zou and Hanjie Chen and Xia Hu},
journal={Transactions on Machine Learning Research},
issn={2835-8856},
year={2025},
url={https://openreview.net/forum?id=HvoG8SxggZ},
note={}
}

@article{chen2024not,
  title={Do not think that much for 2+ 3=? on the overthinking of o1-like llms},
  author={Chen, Xingyu and Xu, Jiahao and Liang, Tian and He, Zhiwei and Pang, Jianhui and Yu, Dian and Song, Linfeng and Liu, Qiuzhi and Zhou, Mengfei and Zhang, Zhuosheng and others},
  journal={arXiv preprint arXiv:2412.21187},
  year={2024}
}

@article{vaswani2017attention,
  title={Attention is all you need},
  author={Vaswani, Ashish and Shazeer, Noam and Parmar, Niki and Uszkoreit, Jakob and Jones, Llion and Gomez, Aidan N and Kaiser, {\L}ukasz and Polosukhin, Illia},
  journal={Advances in neural information processing systems},
  volume={30},
  year={2017}
}

@inproceedings{
prakash2024finetuning,
title={Fine-Tuning Enhances Existing Mechanisms: A Case Study on Entity Tracking},
author={Nikhil Prakash and Tamar Rott Shaham and Tal Haklay and Yonatan Belinkov and David Bau},
booktitle={The Twelfth International Conference on Learning Representations},
year={2024},
url={https://openreview.net/forum?id=8sKcAWOf2D}
}

@article{llama3.2,
title = {Llama 3.2: Revolutionizing edge AI and vision with open, customizable models},
author = {Meta},
year = {2024}
}

@inproceedings{
nikankinarithmetic,
title={Arithmetic Without Algorithms: Language Models Solve Math with a Bag of Heuristics},
author={Yaniv Nikankin and Anja Reusch and Aaron Mueller and Yonatan Belinkov},
booktitle={The Thirteenth International Conference on Learning Representations},
year={2025},
url={https://openreview.net/forum?id=O9YTt26r2P}
}

@inproceedings{
markssparse,
title={Sparse Feature Circuits: Discovering and Editing Interpretable Causal Graphs in Language Models},
author={Samuel Marks and Can Rager and Eric J Michaud and Yonatan Belinkov and David Bau and Aaron Mueller},
booktitle={The Thirteenth International Conference on Learning Representations},
year={2025},
url={https://openreview.net/forum?id=I4e82CIDxv}
}

@inproceedings{dunefskytranscoders,
  title={Transcoders find interpretable LLM feature circuits},
  author={Dunefsky, Jacob and Chlenski, Philippe and Nanda, Neel},
  booktitle={The Thirty-eighth Annual Conference on Neural Information Processing Systems}
}

@article{duttathink,
  title={How to think step-by-step: A mechanistic understanding of chain-of-thought reasoning},
  author={Dutta, Subhabrata and Singh, Joykirat and Chakrabarti, Soumen and Chakraborty, Tanmoy},
  journal={Transactions on Machine Learning Research}
}

@article{yao2024knowledge,
  title={Knowledge circuits in pretrained transformers},
  author={Yao, Yunzhi and Zhang, Ningyu and Xi, Zekun and Wang, Mengru and Xu, Ziwen and Deng, Shumin and Chen, Huajun},
  journal={Advances in Neural Information Processing Systems},
  volume={37},
  pages={118571--118602},
  year={2024}
}

@misc{Caples2025Scaling,
  title = {Scaling Sparse Feature Circuit Finding to Gemma 9B},
  author = {Diego Caples and Jatin Nainani and CallumMcDougall and rrenaud},
  year = {2025},
  url = {https://www.lesswrong.com/posts/PkeB4TLxgaNnSmddg/scaling-sparse-feature-circuit-finding-to-gemma-9b},
}

@article{elhage2022toy,
  title={Toy models of superposition},
  author={Elhage, Nelson and Hume, Tristan and Olsson, Catherine and Schiefer, Nicholas and Henighan, Tom and Kravec, Shauna and Hatfield-Dodds, Zac and Lasenby, Robert and Drain, Dawn and Chen, Carol and others},
  journal={arXiv preprint arXiv:2209.10652},
  year={2022}
}

@article{scherlis2022polysemanticity,
  title={Polysemanticity and capacity in neural networks},
  author={Scherlis, Adam and Sachan, Kshitij and Jermyn, Adam S and Benton, Joe and Shlegeris, Buck},
  journal={arXiv preprint arXiv:2210.01892},
  year={2022}
}

@article{
gurnee2023haystack,
title={Finding Neurons in a Haystack: Case Studies with Sparse Probing},
author={Wes Gurnee and Neel Nanda and Matthew Pauly and Katherine Harvey and Dmitrii Troitskii and Dimitris Bertsimas},
journal={Transactions on Machine Learning Research},
issn={2835-8856},
year={2023},
url={https://openreview.net/forum?id=JYs1R9IMJr},
note={}
}

@article{bricken2023monosemanticity,
       title={Towards Monosemanticity: Decomposing Language Models With Dictionary Learning},
       author={Bricken, Trenton and Templeton, Adly and Batson, Joshua and Chen, Brian and Jermyn, Adam and Conerly, Tom and Turner, Nick and Anil, Cem and Denison, Carson and Askell, Amanda and Lasenby, Robert and Wu, Yifan and Kravec, Shauna and Schiefer, Nicholas and Maxwell, Tim and Joseph, Nicholas and Hatfield-Dodds, Zac and Tamkin, Alex and Nguyen, Karina and McLean, Brayden and Burke, Josiah E and Hume, Tristan and Carter, Shan and Henighan, Tom and Olah, Christopher},
       year={2023},
       journal={Transformer Circuits Thread},
       note={https://transformer-circuits.pub/2023/monosemantic-features/index.html}
    }

@article{llama3.1,
title = {Introducing Llama 3.1: Our most capable models to date},
author = {Meta},
year = {2024}
}

@article{llamascope,
  title={Llama scope: Extracting millions of features from llama-3.1-8b with sparse autoencoders},
  author={He, Zhengfu and Shu, Wentao and Ge, Xuyang and Chen, Lingjie and Wang, Junxuan and Zhou, Yunhua and Liu, Frances and Guo, Qipeng and Huang, Xuanjing and Wu, Zuxuan and others},
  journal={arXiv preprint arXiv:2410.20526},
  year={2024}
}

@article{morris2026learning,
  title={Learning to Reason in 13 Parameters},
  author={Morris, John X and Mireshghallah, Niloofar and Ibrahim, Mark and Mahloujifar, Saeed},
  journal={arXiv preprint arXiv:2602.04118},
  year={2026}
}

@inproceedings{
zhang2025understanding,
title={Understanding and Steering the Cognitive Behaviors of Reasoning Models at Test-Time},
author={Zhenyu Zhang and Xiaoxia Wu and Zhongzhu Zhou and Qingyang Wu and Yineng Zhang and Pragaash Ponnusamy and Harikaran Subbaraj and Jue WANG and Shuaiwen Leon Song and Ben Athiwaratkun},
booktitle={NeurIPS 2025 Workshop on Efficient Reasoning},
year={2025},
url={https://openreview.net/forum?id=yKAEasJpdr}
}

@article{ma2026sparse,
  title={Do Sparse Autoencoders Identify Reasoning Features in Language Models?},
  author={Ma, George and Liang, Zhongyuan and Chen, Irene Y and Sojoudi, Somayeh},
  journal={arXiv preprint arXiv:2601.05679},
  year={2026}
}

@article{li2024optimal,
  title={Optimal ablation for interpretability},
  author={Li, Maximilian and Janson, Lucas},
  journal={Advances in Neural Information Processing Systems},
  volume={37},
  pages={109233--109282},
  year={2024}
}

@article{michel2019sixteen,
  title={Are sixteen heads really better than one?},
  author={Michel, Paul and Levy, Omer and Neubig, Graham},
  journal={Advances in neural information processing systems},
  volume={32},
  year={2019}
}

@inproceedings{voita2019analyzing,
  title={Analyzing multi-head self-attention: Specialized heads do the heavy lifting, the rest can be pruned},
  author={Voita, Elena and Talbot, David and Moiseev, Fedor and Sennrich, Rico and Titov, Ivan},
  booktitle={Proceedings of the 57th annual meeting of the association for computational linguistics},
  pages={5797--5808},
  year={2019}
}

@inproceedings{teplica-etal-2025-sciurus,
    title = "{SCIUR}us: Shared Circuits for Interpretable Uncertainty Representations in Language Models",
    author = "Teplica, Carter  and
      Liu, Yixin  and
      Cohan, Arman  and
      Rudner, Tim G. J.",
    editor = "Chiruzzo, Luis  and
      Ritter, Alan  and
      Wang, Lu",
    booktitle = "Proceedings of the 2025 Conference of the Nations of the Americas Chapter of the Association for Computational Linguistics: Human Language Technologies (Volume 1: Long Papers)",
    month = apr,
    year = "2025",
    address = "Albuquerque, New Mexico",
    publisher = "Association for Computational Linguistics",
    url = "https://aclanthology.org/2025.naacl-long.618/",
    doi = "10.18653/v1/2025.naacl-long.618",
    pages = "12451--12469",
    ISBN = "979-8-89176-189-6"
}

@inproceedings{
park2026asguard,
title={{ASG}uard: Activation-Scaling Guard to Mitigate Targeted Jailbreaking Attack},
author={Yein Park and Jungwoo Park and Jaewoo Kang},
booktitle={The Fourteenth International Conference on Learning Representations},
year={2026},
url={https://openreview.net/forum?id=wmiEXNEXPs}
}

@inproceedings{
yin2025which,
title={Which Attention Heads Matter for In-Context Learning?},
author={Kayo Yin and Jacob Steinhardt},
booktitle={Forty-second International Conference on Machine Learning},
year={2025},
url={https://openreview.net/forum?id=C7XmEByCFv}
}
\bibliographystyle{colm2026_conference}
\clearpage
\appendix
\section{Appendix}
\subsection{Related Work}

\subsubsection{Supervised Fine-Tuning (SFT) \& Distillation}
Post-training is a crucial stage that adapts a general-purpose pretrained LLM for specialized tasks such as complex reasoning~\citep{rl_survey}.
Supervised Fine-Tuning (SFT) adapts a pretrained model to a specific tasks by training it on a curated dataset of input-output examples~\citep{wei2022finetuned}.
In the context of reasoning, a powerful technique is to use a large, more capable "teacher" model (e.g., DeepSeek-R1~\citep{r1}) to generate high-quality, step-by-step reasoning races, often called Chain-of-Tought (CoT)~\citep{cot} prompts.
A smaller "student" model is then fine-tuned on this synthetic dataset, learning to mimic the teacher's reasoning process~\citep{kard}.
SFT forces the student model's output distribution to match the teacher's, and this direct and forceful adaptation often results in significant, dense updates to the model's parameter by memorizing specific reasoning paths~\citep{chu2025sft}.
This form of knowledge distillation has proven effective for creating capable open-source reasoning models~\citep{toshniwal2024openmathinstruct}.
Here, we utilize distilled version of DeepSeek-R1 for the corresponding Qwen2.5 Math~\citep{qwen2.5math}, and do SFT with sampled OpenR1-Math-220k dataset for comparison~\citep{openr1}.

\subsubsection{Reinforcement Learning with Verifiable Rewards (RLVR)}

Reinforcement learning (RL) offers an alternative paradigm where a model learns by interacting with an environment and receiving reward signals~\citep{ouyang2022training}.
It is particularly well-suited for tasks like the mathematical reasoning where the correctness of a final answer can be automatically verified, providing a clear, albeit sparse, reward signal.
This Reinforcement Learning with Verifiable Rewards (RLVR) allows the model to explore different reasoning paths and reinforces those that lead to correct outcomes, without being constrained to a signal gold path as in SFT.
Recent evidence from~\citep{morris2026learning} suggests that RLVR may improve performance less by injecting substantial new knowledge than by selectively activating or reconfiguring capabilities already latent in the base model.
A prominent RL algorithm used for training reasoning models is Group Relative Policy Optimization (GRPO)~\citep{shao2024deepseekmath}, a variant of Proximal Policy Optimization (PPO)~\citep{ppo}, designed to be more memory efficient and stable training.
We adopt GRPO to implement RLVR for mathematical reasoning; the full objective and training formulation are detailed in~\S\ref{app:detail_grpo}.

\subsubsection{Mechanistic Interpretability for LLM Reasoning}
\label{subsec:mech-inter}

Mechanistic interpretability seeks to explain model behavior via internal mechanisms, and one common approach studies small, causally meaningful ``circuits" connecting attention heads and MLPs~\citep{circuitgrokking,conmy2023towards,ameisen2025circuit,lindsey2025biology}. 
Circuits have been reverse-engineered for indirect object identification in GPT-2 Small~\citep{wang2023interpretability}, for factual and temporal knowledge~\citep{yao2024knowledge,temporalhead}, and for chain-of-thought reasoning~\citep{duttathink,cabannes2024iteration}, while arithmetic work shows that models rely on a ``bag of heuristics" implemented by sparse MLP features rather than a single clean algorithm~\citep{nikankinarithmetic}.
Complementary head-level analyses of reasoning models have identified specialized attention heads associated with behaviors such as verification and backtracking, and shown that lightly intervening on these heads can steer reasoning at inference time~\citep{zhang2025understanding}.
However, interpreting such circuits at the level of individual units is complicated by \emph{polysemanticity}: superposition makes neurons and heads mix multiple unrelated features, and many human-interpretable features appear only as sparse combinations of neurons rather than clean single units~\citep{elhage2022toy,scherlis2022polysemanticity,gurnee2023haystack}. 

This has motivated feature-based approaches such as Sparse Feature Circuits and their large-scale extensions~\citep{markssparse,Caples2025Scaling} and Transcoder-based MLP replacements~\citep{dunefskytranscoders}, which learn sparse latent features for more precise circuit editing but require substantial extra training and are currently implemented for only a few architectures.
At the same time, recent work cautions that many contrastively selected SAE ``reasoning features” may reflect lexical or discourse correlates of chain-of-thought rather than reasoning computations themselves, underscoring the need for falsification in feature-level analyses of reasoning~\citep{ma2026sparse}.
Head- and neuron-level circuit analyses nonetheless remain the default abstraction in transformer-circuits work and continue to yield experimentally testable insights~\citep{wang2023interpretability,yao2024knowledge,temporalhead}, so we adopt this conventional perspective and operate directly on native attention heads.
By avoiding per-layer sparse autoencoders or transcoders, our analysis is much more computationally efficient and easily transferable across architectures and post-training regimes, at the cost of some residual polysemanticity. 
Most closely related to our goals,~\cite{prakash2024finetuning} find that fine-tuning on entity tracking mainly strengthens existing mechanisms rather than creating new ones, whereas in our math-only SFT and GRPO setting with an explicit \texttt{<think>} token we observe emergent ``reasoning heads" that are negligible in the base model but become critical after post-training, suggesting that circuit reorganisation depends strongly on both task domain and training paradigm.

\subsection{Detail of EAP-IG Calculation}
\label{subsec:eapig-detail}
\textbf{Global path.} 
The IG path is defined over the \emph{entire token-embedding sequence}: 
we linearly interpolate between corrupted and clean inputs as \(z' + \alpha (z - z')\) with \(\alpha=\tfrac{k}{m}\), \(k=1,\dots,m\). No pooling into a single “document embedding” is used.

\textbf{Input of \(v\).} 
For a node \(v\) (attention head block or MLP), the “input of \(v\)” is the \emph{residual-stream pre-activation} that \(v\) receives at its destination positions, i.e., the sum of all parents’ outputs just before \(v\) applies its operation. 
Accordingly, the gradient in \eqref{eq:eap-ig} is \(\nabla_{z_v}\mathcal L\) with respect to that residual vector.

\textbf{Token granularity and per-example score.} 
While the path lives in sequence space, the edge score for $(u\!\to\!v)$ is evaluated at coordinates corresponding to $(v)$'s destination positions. 
For next-token objectives we use the position $(t)$ whose logits are evaluated; for sequence-level objectives we average over supervised positions $(T^*)$. The per-example score is
\begin{equation}
\mathrm{score}(u\!\to\!v \mid x)
=
\left\langle
\Delta z_u(x),\;
\frac{1}{m}\sum_{k=1}^{m}
\big(\nabla_{z_v}\mathcal L\big)\Big|_{\,z' + \frac{k}{m}(z-z')}
\right\rangle,
\end{equation}
where \(\langle\cdot,\cdot\rangle\) denotes the dot product in the residual dimension.

\textbf{Aggregation and selection.} 
We rank edges using a dataset aggregate, e.g.,
\(\mathbb{E}_{x}\!\left[\lvert \mathrm{score}(u\!\to\!v \mid x)\rvert\right]\).
Using \(\Delta z_u=z_u-z'_u\) or \(z'_u-z_u\) only flips the sign; 
absolute aggregation makes ranking invariant. 
We select top-\(n\) edges, prune isolated nodes, and validate faithfulness by ablating all non-circuit edges.

\textbf{Practical choices.} 
We typically use \(m\in[5,8]\) Riemann steps and a task-agnostic divergence (e.g., KL) computed at the same evaluation positions as above; 
rankings are robust without extra normalization, though optional rescaling can be applied for cross-model comparability.

\subsection{Detail of Sparse Feature Circuit Analysis}
\label{subsec:sfc-detail}
\textbf{Construction of Graph.} 
Constructing full Sparse Feature Circuits~\citep{markssparse} implies a prohibitive computational cost, scaling with the number of training methods, model checkpoints, layers, and components. 
To make this tractable while leveraging the disentanglement benefits of Sparse Autoencoders (SAEs)~\citep{bricken2023monosemanticity}, we limit our scope to a direct comparison between Llama-3.1-8B (Base)~\citep{llama3.1} and DeepSeek-R1-Distill-Llama-8B~\citep{r1}, where both model's full SAEs for residual stream are available through Neuronpedia~\citep{llamascope}.
We utilize those pre-trained Residual Stream SAEs to decompose residual activations into sparse features $f \in \mathbb{R}^{d_{\text{SAE}}}$. 
However, for Attention and MLP blocks where SAE training is computationally demanding, we retain a dense representation using \emph{identity dictionaries}, and compute attribution scores for sparse features in the residual stream and for dense block outputs in the Attention and MLP layers using the same mathematical algorithm as in EAP-IG~\citep{eapig}. 
Input dataset is same with previous EAP-IG analysis, which is AIME base prompt with sampled answer.

\textbf{Aggregated Importance and Shift Measurement.} 
Since the learned dictionary bases of SAEs differ between the base and post-trained models, a direct feature-to-feature comparison is infeasible. 
Instead, we aggregate importance at the component level to quantify macroscopic shifts. 
For a model $M$, layer $\ell$, and component $c \in \{\textsc{Resid}, \textsc{Attn}, \textsc{Mlp}\}$, the importance $I_M(\ell, c)$ is the sum of absolute attribution scores of all constituent nodes (active SAE features for Resid, or the dense block for Attn/Mlp). 
We then visualize the shift using the symmetric relative difference defined in \S\ref{app:detail_effect}:
\begin{equation}
    E(\ell, c) = \frac{\hat{I}_{M_{\text{post}}}(\ell, c) - \hat{I}_{M_{\text{pre}}}(\ell, c)}{\hat{I}_{M_{\text{post}}}(\ell, c) + \hat{I}_{M_{\text{pre}}}(\ell, c) + \varepsilon},
\end{equation}
where $\hat{I}$ denotes the globally normalized importance. This metric highlights which computational stages become more critical after distillation.

\begin{figure}[t]
\vspace{-10pt}
\begin{center}
\includegraphics[width=0.52\columnwidth]{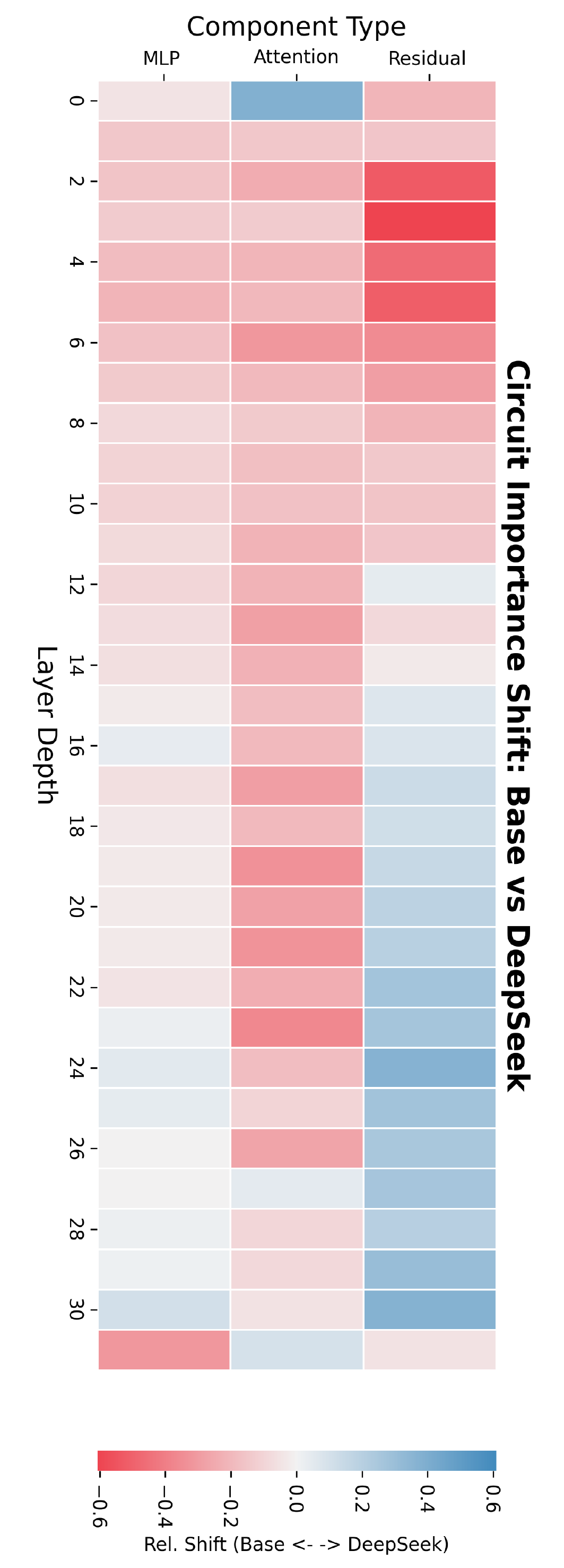}
\end{center}%
\vspace{-10pt}
\caption{
Component-level importance shift between Llama-3.1-8B (Base) and DeepSeek-R1-Distill-Llama-8B derived from Sparse Feature Circuits. 
Columns represent the aggregated attribution score for MLP, Attention, and Residual components across layers.
The color encodes the symmetric effect measure.
Blue (positive) indicates components where the DeepSeek model places higher causal weight (e.g., Layer 0 Attention and late-stage Residual streams), while Red (negative) indicates components more dominant in the Base model.}
\label{fig:sparse_feature_circuit}
\vspace{-10pt}
\end{figure}

\textbf{Results and Discussion.} 
The analysis reveals distinct patterns in computational reallocation. 
Figure~\ref{fig:sparse_feature_circuit} shows component-level importance with a single heatmap.
Consistent with our head-level EAP-IG findings, we observe a strong emergence of importance in \textbf{Layer 0 Attention}, suggesting early-stage emergence of attention heads remains crucial. 
Notably, the \textbf{Residual Stream} features exhibit a progressive strengthening in the mid-to-late layers, indicating a reliance on deep, disentangled representations for reasoning. 
The \textbf{MLP} blocks also show increased importance in later layers, albeit less dominantly than residuals. 
While this SAE-based approach offers reduced polysemanticity and corroborates our main findings, its coarse granularity at the Attention/MLP block level prevents the precise identification of specialized heads.
Therefore, given the trade-off between feature interpretability from enormous computational cost and practical granular component tracking, we retain the standard head-level EAP-IG as our primary analytical framework.

\subsection{Experimental Setup}
\label{subsec:exp_setup}
\paragraph{Models.}
We select a consistent family of models to serve as the testbed for our analysis among similar architecture and design.
As Qwen series make it possible to compare almost every possible reasoning training, we specifically pick this model variations and analyze deeply.
The models include:
\begin{itemize}
    \item Baseline Models: Qwen2.5-Math-1.5B-Instruct and Qwen2.5-Math-7B-Instruct~\citep{qwen2.5math}, which are strong base models pretrained with a focus on mathematical capabilities.
    \item Distilled Models: DeepSeek-R1-Distill-Qwen-1.5B and 7B~\citep{r1}, which represent the outcome of knowledge distillation from a powerful teacher reasoning model.
    \item Think On/Off Model: Qwen3-8B~\citep{qwen3}, which features a Think On/Off capability across various open source models, allowing for controlled study of selective reasoning activation.
\end{itemize}
We additionally adopt Llama-3.2-1B-Instruct~\citep{llama3.2} for generalizable re-implementation, though it cannot be compared with the corresponding DeepSeek distillation and think on/off model as they do not exist.
\vspace{-5pt}
\paragraph{Datasets.}
Our training and evaluation cover the well-established, widely-used reasoning datasets:
\begin{itemize}
    \item Training: For SFT and GRPO, we utilize standard, large-scale reasoning datasets, including OpenR1-Math-220k~\citep{openr1} and GSM8K~\citep{gsm8k}, which contain a diverse set of mathematical problems and their solutions.
    \item Evaluation: To assess both in-domain and out-of-domain generalization, we employed a comprehensive suite of benchmarks: AIME'24 and AIME'25 (American Invitational Mathematics Examination)~\citep{aime}, AMC (American Mathematics Competitions)~\citep{AMC2023}, GPQA (Graduate-Level Google-Proof Q\&A)~\citep{gpqa}, MATH-500~\citep{lightman2024lets} and TriviaQA~\citep{triviaqa} for general knowledge.
\end{itemize}

\begin{figure}[t]
\begin{center}
\includegraphics[width=\columnwidth]{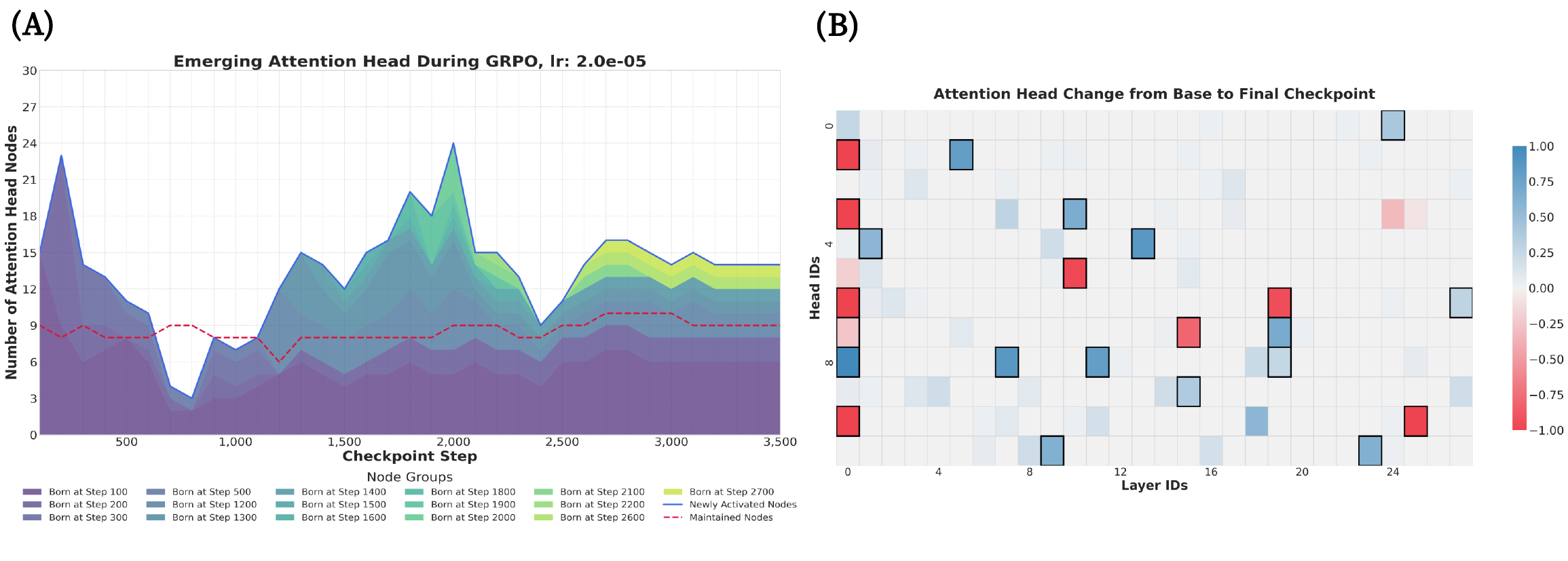}
\end{center}%
\vspace{-10pt}
\caption{Emergent attention heads in Qwen2.5-Math-1.5B during GRPO on OpenR1-Math-220k~\citep{openr1} and learning rate 2e-05, with circuits constructed on~\cite{aime}.
The figure follows the same visualization protocol as Figure~\ref{fig:main_fig_sft}.
(A) Cohort analysis across checkpoints; blue curve for newly activated heads, red dashed curve for retained base-model heads, and stacked areas group heads by emergence timing.
(B) Activation-frequency heatmap with the same color convention as Figure~\ref{fig:main_fig_sft} (red for base-model, blue for emergent heads); heads active at the final checkpoint are outlined in black.
}
\label{fig:main_fig_grpo_math220_lr2}
\end{figure}

\paragraph{Training \& Evaluation.}
For each post-training method, we follow established best practices and maintain consistent hyperparameters where possible to facilitate fair comparison. 
For GRPO, we train a Qwen2.5-Math-1.5B-Instruct for 3 epochs, saving checkpoints every 100 steps to enable a temporal analysis of circuit formation.
For SFT, we used a setup designed to mirror the GRPO training process in terms of data exposure.
We also utilize Light-R1~\citep{lightr1} as our codebase, modifying it so that the pass@1 evaluation metric is computed as the average over multiple responses for each setting.
All training and inference are done with two NVIDIA H100 GPUs(80GB).
Hyper-parameter setup for each post-training is like below:
\begin{itemize}
    \item SFT~\citep{wei2022finetuned}: 
    learning rate $4.0e-5$, 5 training epochs, 100 steps for saving and circuit construction, Bfloat16, warm-up ratio $0.03$.
    For Llama3.2 1B: learning rate $4.0e-5$, 5 training epochs, 100 steps for circuit construction, Bfloat16, warm-up ratio $0.03$
    \item GRPO~\citep{shao2024deepseekmath} with OpenrR1-Math-220k:
    learning rate $1.0e-6$ for main result and $2.0e-5$ for comparison in Figure~\ref{fig:main_fig_grpo_math220_lr2}, 3 training epochs, 100 steps for saving and circuit construction, Bfloat16, warm-up ratio $0.1$, reward\_weights $1.0$, 16 generations.
    For Llama3.2 1B: learning rate $2.0e-7$, 3 training epochs, 100 steps for saving and circuit construction, Bfloat16, warm-up ratio $0.1$, reward\_weights $1.0$, 16 generations.
    \item GRPO~\citep{shao2024deepseekmath} with GSM8K:
    learning rate $5e-6$, 1 training epoch, 100 steps for saving and circuit construction, Bfloat16, warm-up ratio $0.1$, reward\_weights $1.0$, 16 generations.
\end{itemize}

For the system prompt of GRPO training, we use basic recipes of OpenR1~\citep{openr1}.
\begin{tcolorbox}[colback=gray!3,colframe=black]
You are a helpful AI Assistant that provides well-reasoned and detailed responses. You first think about the reasoning process as an internal monologue and then provide the user with the answer. Respond in the following format: \\
\verb|<think>\n...\n</think>\n<answer>\n...\n</answer>|
\end{tcolorbox}

\subsection{Circuit Construction Setup}
\label{subsec:circuit_setup}
\begin{figure}[t]
\vspace{-10pt}
\begin{center}
\includegraphics[width=\columnwidth]{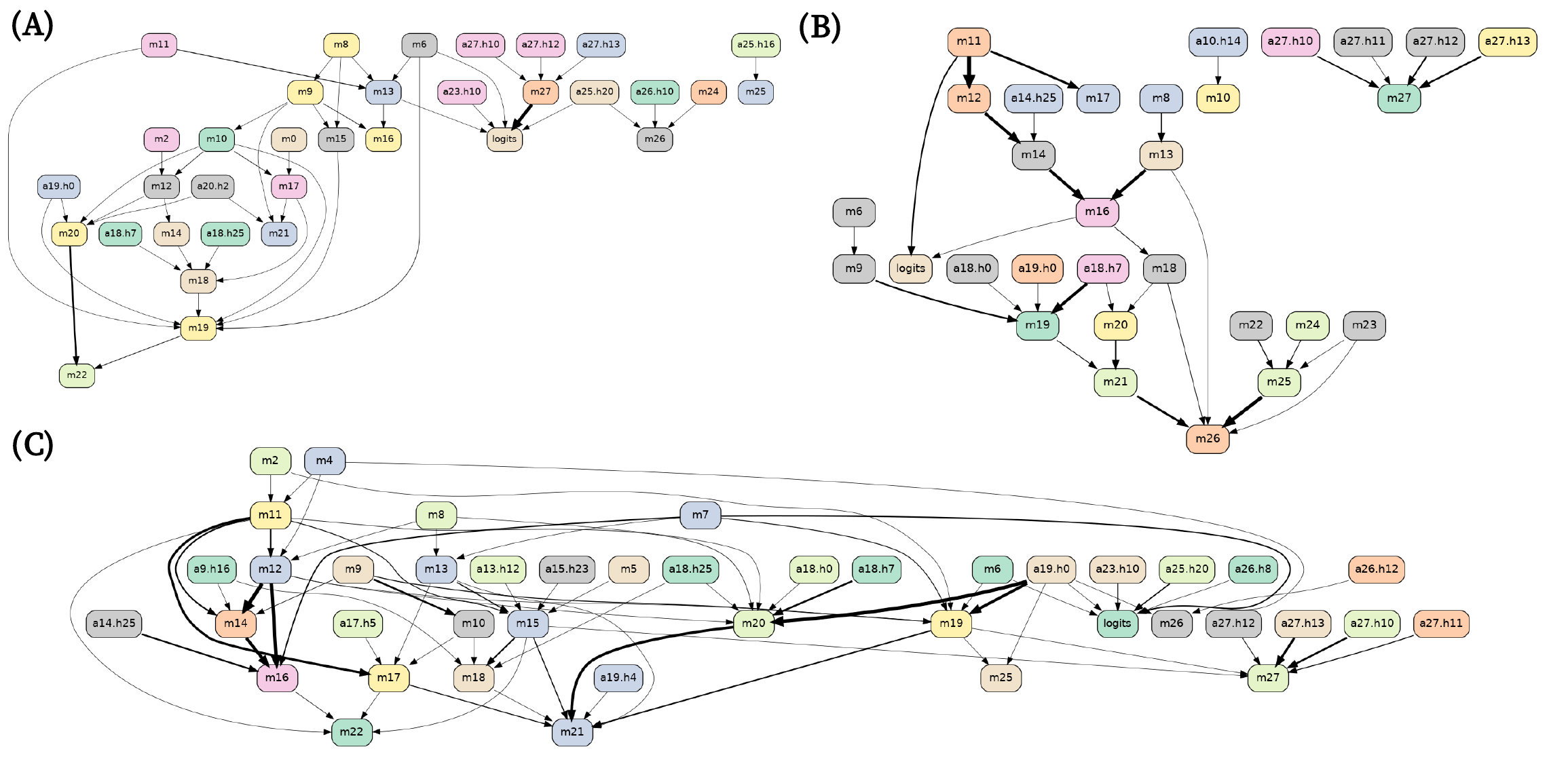}
\end{center}%
\vspace{-10pt}
\caption{Actual Example of Circuits.
Color of nodes are randomly mapped to differentiate each others.
(A) denotes AIME circuit with baseline model, Qwen-2.5-Math-7B.
(B) shows AIME circuit with DeepSeek-R1-Distill-Qwen-7B.
(C) is the comparative example with same AIME dataset, which is constructed with DeekSeek-R1-Distill-Qwen-7B and its own sampled answer, without explicit \texttt{<think>}.
(C) is more complex than other two circuits, which could be mixed with confusable attention heads.
The trend of this enormous attention heads in (C) is also similar with the thinking off mode in Figure~\ref{fig:circuit_example_thinkonoff} (B), where the model compensate its performance gap through large emergent attention heads.
}
\label{fig:circuit_example}
\vspace{-10pt}
\end{figure}

\begin{figure}[t]
\vspace{-10pt}
\begin{center}
\includegraphics[width=\columnwidth]{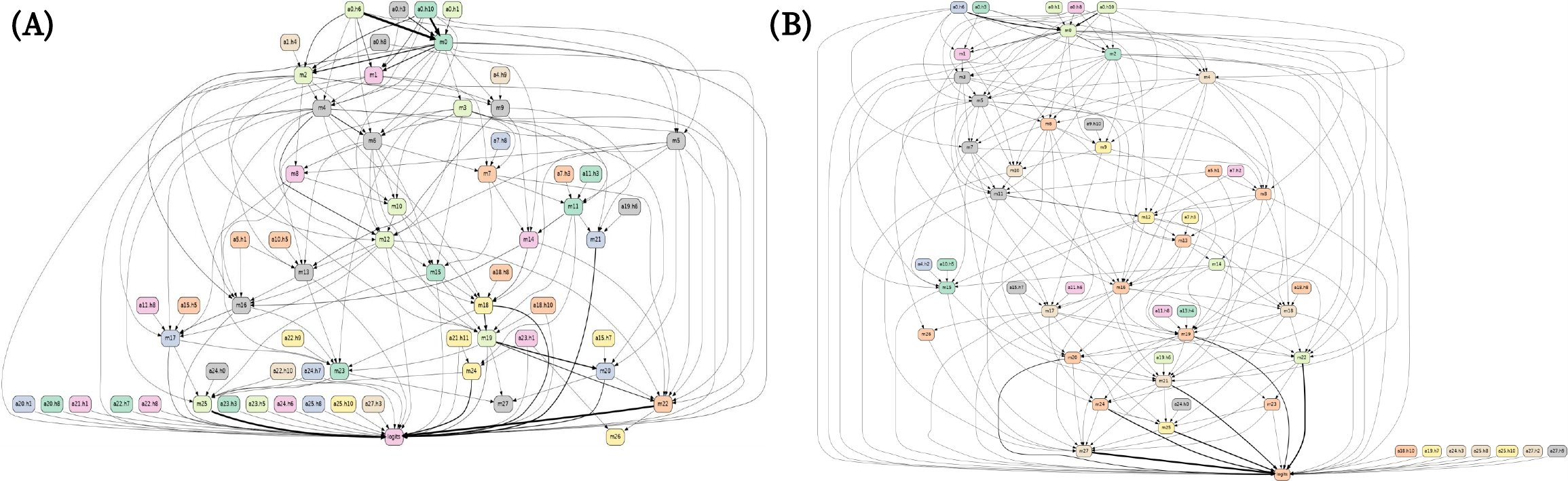}
\end{center}%
\vspace{-10pt}
\caption{Actual Example of Circuits After Post-Training.
Color of nodes are randomly mapped to differentiate each others.
(A) denotes AIME circuit after SFT with baseline model, Qwen-2.5-Math-1.5B.
(B) shows AIME circuit after GRPO with the same baseline model.
(A) activates more attention heads while (B) has more complexly connected specific nodes which refer its internalized high-level mathematical reasoning.
}
\label{fig:circuit_example_post_training}
\vspace{-10pt}
\end{figure}

\begin{figure}[t]
\vspace{-10pt}
\begin{center}
\includegraphics[width=\columnwidth]{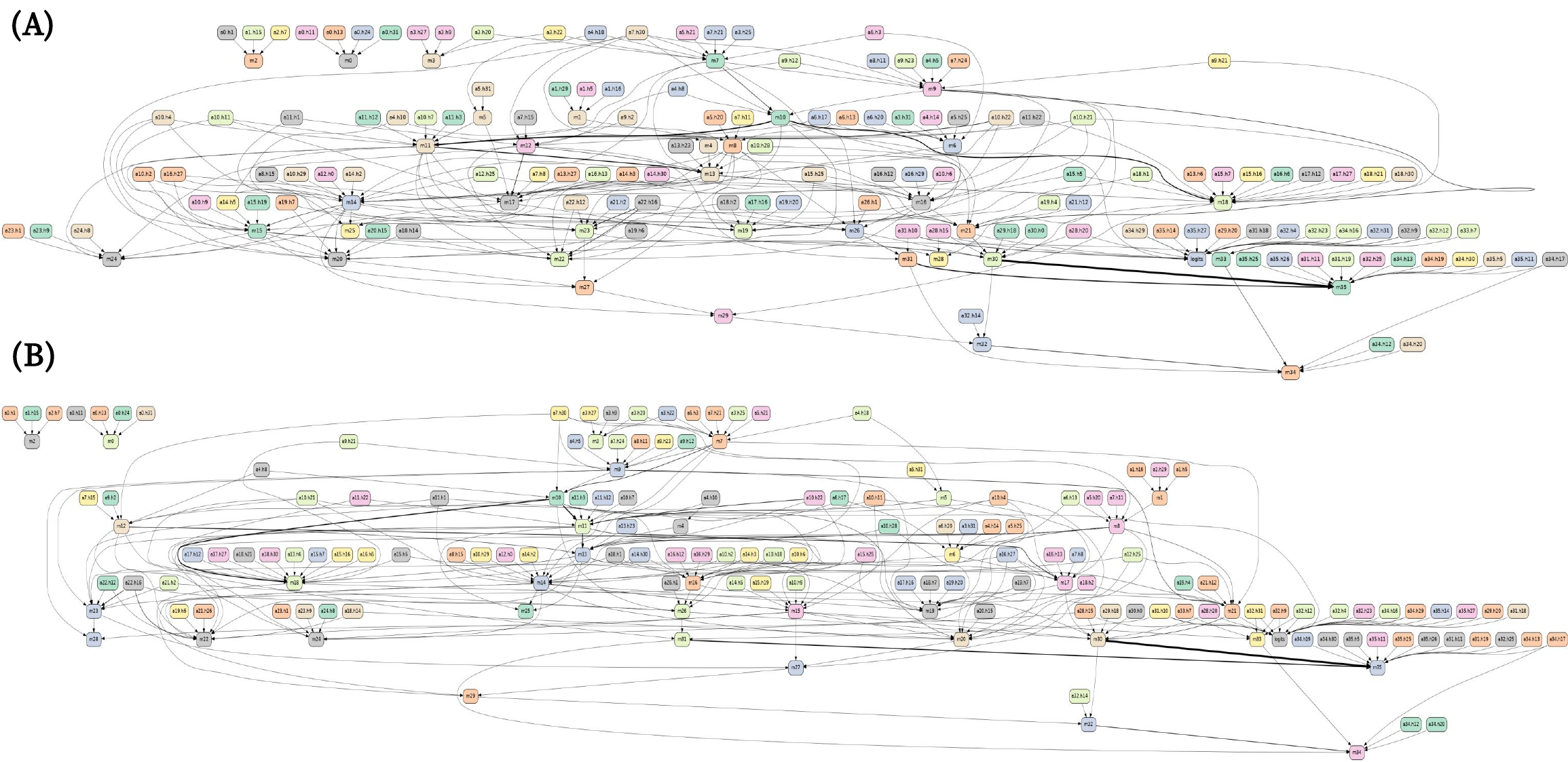}
\end{center}%
\vspace{-10pt}
\caption{Actual Example of Circuits of Think On/Off.
Color of nodes are randomly mapped to differentiate each others.
(A) denotes AIME circuit of Thinking on mode in Qwen3-8B.
(B) shows AIME circuit of Thinking off on the same baseline model.
(B) activates more attention heads, in contrast, (A) has more complexly connected specific nodes which refer its internalized high-level mathematical reasoning, similar as GRPO circuit in Figure~\ref{fig:circuit_example_post_training} (B).
}
\label{fig:circuit_example_thinkonoff}
\vspace{-10pt}
\end{figure}

\begin{figure}[t]
\vspace{-10pt}
\begin{center}
\includegraphics[width=\columnwidth]{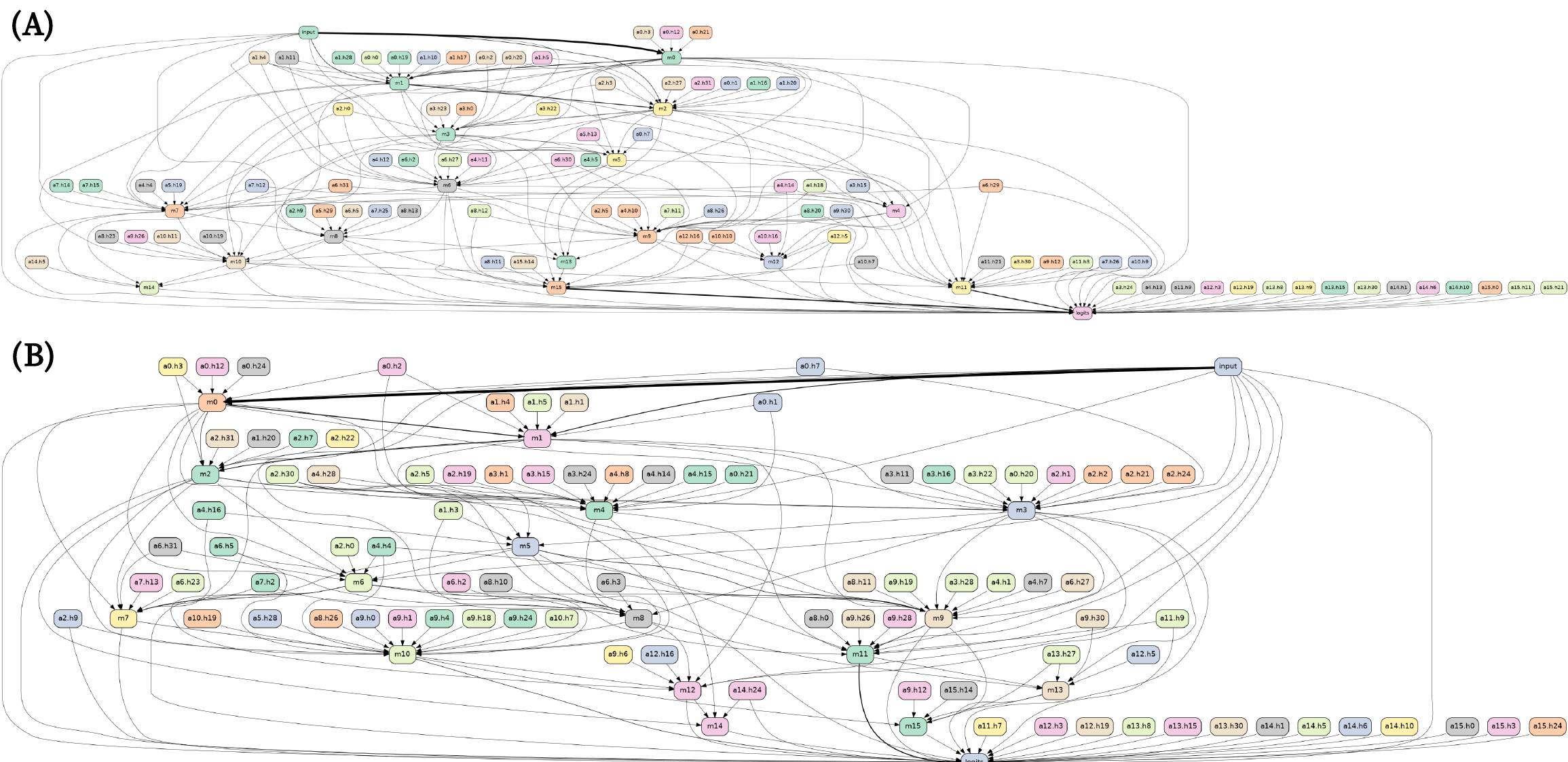}
\end{center}%
\vspace{-10pt}
\caption{Actual Example of Circuits of Llama-3.2-1B-Instruct.
Color of nodes are randomly mapped to differentiate each others.
(A) denotes AIME circuit of Llama 3.2 after SFT with OpenR1-Math-220k dataset.
(B) shows AIME circuit of Llama 3.2 after GRPO with OpenR1-Math-220k dataset.}
\label{fig:circuit_example_llama}
\vspace{-10pt}
\end{figure}

We construct circuits using EAP-IG~\citep{eapig}, where \emph{ig-step} is 100 and \emph{top-\(n\)} is 5000.
We also simplify each circuits with the threshold $\tau = 0.1$ for filtering out important edges and nodes.
Examples of simplified circuits among various models are in Figure~\ref{fig:circuit_example},~\ref{fig:circuit_example_post_training}, and~\ref{fig:circuit_example_thinkonoff}.
Figure~\ref{fig:circuit_example_llama} is the examples of simplified circuits with Llama3.2 1B.
\paragraph{Prompt Settings.}
We sample various responses of baseline models and reasoning models, then make an input prompt for circuit construction using chat template.
\begin{tcolorbox}[
      enhanced, breakable,
      colback=gray!3, colframe=black,
      colbacktitle=black, coltitle=white,
      fonttitle=\bfseries, title=Reasoning Model,
      segmentation style={dashed} 
    ]
    \texttt{<think>}Okay, so I have this problem where Aya goes ...
    \tcbline
    \texttt{<think>}Alright, so I have this geometry problem here ...
    \tcbline
    \texttt{<think>}Okay, so I need to find the eigenvector ...
    \tcbline
    \texttt{<think>}...
\end{tcolorbox}
\begin{tcolorbox}[
      enhanced, breakable,
      colback=gray!3, colframe=black,
      colbacktitle=black, coltitle=white,
      fonttitle=\bfseries, title=Baseline model,
      segmentation style={dashed} 
    ]
    We'll use Python to help us ...
    \tcbline
    To determine the molecular
\end{tcolorbox}
For Llama3.2 1B, we sample responses of baseline models and after reasoning to construct circuits.
\begin{tcolorbox}[
      enhanced, breakable,
      colback=gray!3, colframe=black,
      colbacktitle=black, coltitle=white,
      fonttitle=\bfseries, title=Models After Post Training,
      segmentation style={dashed} 
    ]
    \texttt{<think>}Step 1: Define the variables and given conditions Let's denote ...
\end{tcolorbox}
\begin{tcolorbox}[
      enhanced, breakable,
      colback=gray!3, colframe=black,
      colbacktitle=black, coltitle=white,
      fonttitle=\bfseries, title=Models Before Post Training,
      segmentation style={dashed} 
    ]
    To solve this problem, we'll ...
\end{tcolorbox}

\subsection{Detail of Effect and Importance Measure}
\label{app:detail_effect}
Our effect analysis reuses the EAP-IG edge scores already computed for circuit extraction (\S~\ref{subsec:circuit-analysis}).  
For a given model $M$ and input $x$ from a benchmark dataset $\mathcal{D}$, EAP-IG assigns to each edge $(u \to v)$ in the circuit $C^{(M)}(x)$ a scalar attribution score $s^{(M)}_x(u \to v) \in \mathbb{R}$, which we obtain after thresholding on $\lvert s^{(M)}_x(u \to v)\rvert$ to keep only top-attribution edges.
We treat attention heads as modules and aggregate edge-level scores into a head-level importance matrix.

\paragraph{Head-level importance.}
Let $a_{\ell,h}$ denote the attention head at layer $\ell$ and index $h$.
For model $M$, we define the (unnormalized) importance of $a_{\ell,h}$ as the sum of absolute EAP-IG scores over all circuits and all edges whose source node is that head:
\begin{equation}
    \tilde I_M(\ell,h)
    \;=\;
    \sum_{x \in \mathcal{D}}
    \sum_{\substack{(u \to v) \in C^{(M)}(x)\\ u = a_{\ell,h}}}
    \bigl\lvert s^{(M)}_x(u \to v) \bigr\rvert.
\end{equation}
To allow comparison across models, we apply a global normalization so that the total mass of importance is $1$:
\begin{equation}
    I_M(\ell,h)
    \;=\;
    \frac{\tilde I_M(\ell,h)}
         {\displaystyle \sum_{\ell',h'} \tilde I_M(\ell',h')}.
\end{equation}
This yields a head-level importance matrix $I_M \in \mathbb{R}_{\ge 0}^{L \times H}$, where $L$ is the number of layers and $H$ the number of heads per layer.

\paragraph{Effect measure between pre- and post-trained models.}
Given a pre-trained (base) model $M_{\text{pre}}$ and a post-trained model $M_{\text{post}}$ (e.g., DeepSeek-distilled, SFT, or GRPO-trained), both evaluated on the same dataset $\mathcal{D}$ with identical EAP-IG hyperparameters and edge-thresholding, we quantify the change in importance of head $(\ell,h)$ by the symmetric effect measure
\begin{equation}
    E(\ell,h)
    \;=\;
    \frac{ I_{M_{\text{post}}}(\ell,h) - I_{M_{\text{pre}}}(\ell,h) }
         { I_{M_{\text{post}}}(\ell,h) + I_{M_{\text{pre}}}(\ell,h) + \varepsilon },
    \label{eq:effect}
\end{equation}
where $\varepsilon > 0$ is a small constant (we use $\varepsilon = 10^{-6}$) to avoid division by zero.
By construction, $E(\ell,h) \in [-1,1]$, with positive values indicating increased attribution-based importance of $a_{\ell,h}$ in the post-trained model and negative values indicating decreased importance.

\begin{figure}[t]
\vspace{-10pt}
\begin{center}
\includegraphics[width=\columnwidth]{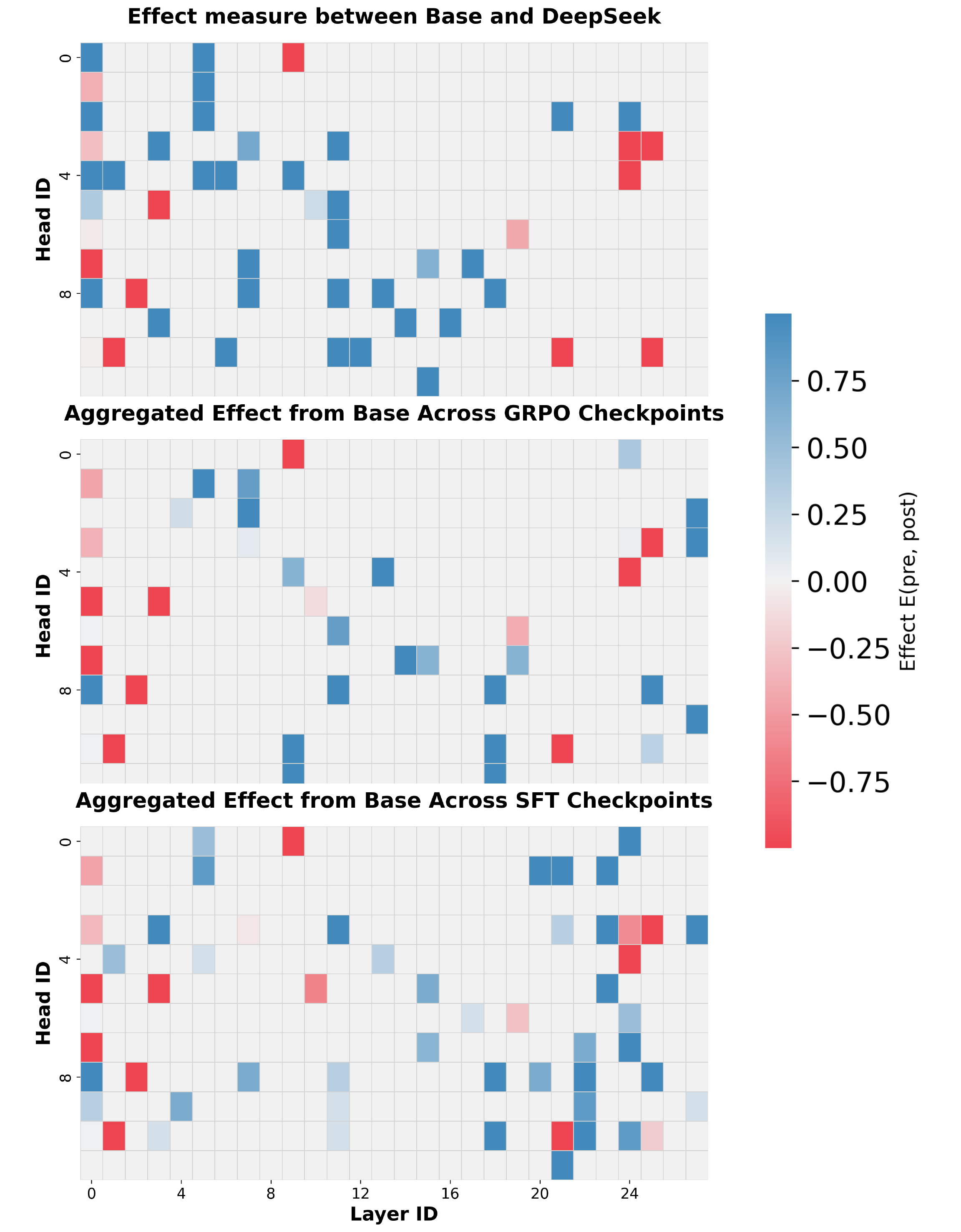}
\end{center}%
\vspace{-10pt}
\caption{
Head-level effect maps for Qwen2.5-Math-1.5B and its post-trained variants.
From top to bottom: 
Effect between the base Qwen2.5-Math-1.5B model and the DeepSeek-distilled reasoning model;
Effect aggregated across GRPO checkpoints (500-step intervals from 500 to 2500 steps) trained from the same base;
Effect aggregated across SFT checkpoints (200-step intervals).
Each cell corresponds to an attention head $(\ell,h)$, and the color encodes the symmetric effect measure
$E(\ell,h) = \bigl(I_{\text{post}}(\ell,h) - I_{\text{pre}}(\ell,h)\bigr)\big/\bigl(I_{\text{post}}(\ell,h) + I_{\text{pre}}(\ell,h) + \varepsilon\bigr)$,
where $I_{\text{pre}}$ and $I_{\text{post}}$ are the EAP-IG–based head importances defined in \S\ref{app:detail_effect}.
Blue (red) indicates increased (decreased) attribution-based importance of the head relative to the base model.
The high-magnitude heads in these maps qualitatively align with the high-frequency circuit heads in Figure~\ref{fig:main_fig_sft} (B) and \ref{fig:main_fig_grpo_math220} (C), indicating that our frequency-based circuit analysis is consistent with the attribution-based importance view.}
\label{fig:circuit_effect}
\vspace{-10pt}
\end{figure}

For training regimes with multiple checkpoints $\ M_{\text{post}}^{(t)}\}_{t\in\mathcal{T}}$ (e.g., SFT or GRPO), we compute \eqref{eq:effect} for each checkpoint $t$ to obtain $E^{(t)}(\ell,h)$ and then aggregate along the time axis via a simple arithmetic mean:
\begin{equation}
    \bar E(\ell,h)
    \;=\;
    \frac{1}{\lvert \mathcal{T} \rvert}
    \sum_{t \in \mathcal{T}} E^{(t)}(\ell,h).
\end{equation}
The resulting matrix $\bar E \in [-1,1]^{L \times H}$ is visualized as the effect heatmaps in Figure~\ref{fig:circuit_effect}, where blue (red) cells correspond to heads whose EAP-IG importance increases (decreases) relative to the base model.
Note that, because circuits are defined using a fixed attribution threshold, these measures capture importance reallocation within the \emph{top-attribution circuits} considered in our analysis.

\subsection{Detail of Ablation Setup and Interpretation}

\label{app:zero-vs-mean-ablation}
Our ablation analysis is designed to answer a behavioral necessity question:
given a set of attention heads $H$ identified from our circuit analysis, how task-level benchmark performance changes when their contribution is removed?
This objective differs from minimal circuit reconstruction under a tightly controlled prompt distribution, where distribution-faithful interventions such as mean ablation or counterfactual ablation are often preferable~\citep{wang2023interpretability,prakash2024finetuning,li2024optimal}.
Accordingly, our intervention is intended as a head-removal sensitivity test rather than as a claim that zero ablation is the uniquely correct operator for mechanistic effect estimation.

\paragraph{Formalization.}
Let $f$ denote the intact model, and let
$a_{h,t}(x)\in\mathbb{R}^{d_{\mathrm{model}}}$ be the vector written by attention head $h$
at token position $t$ into the residual stream for input $x$.
For an ablated head set $H$, we define the modified head output under zero ablation as
\begin{equation}
\tilde{a}^{\mathrm{zero}}_{h,t}(x)=
\begin{cases}
0, & h\in H,\\
a_{h,t}(x), & h\notin H.
\end{cases}
\label{eq:zero-ablation}
\end{equation}
That is, the entire write vector of the ablated head is removed from the residual stream.

By contrast, under mean ablation, the ablated head is replaced by a reference mean:
\begin{equation}
\tilde{a}^{\mathrm{mean}}_{h,t}(x)=
\begin{cases}
\mu^{\mathrm{ref}}_{h,t}, & h\in H,\\
a_{h,t}(x), & h\notin H,
\end{cases}
\qquad
\mu^{\mathrm{ref}}_{h,t}
=
\mathbb{E}_{x'\sim\mathcal{D}_{\mathrm{ref}}}
\left[a_{h,t}(x')\right],
\label{eq:mean-ablation}
\end{equation}
where $\mathcal{D}_{\mathrm{ref}}$ is a reference distribution, ideally matched in token
position and prompt/template structure~\citep{wang2023interpretability}.

Let $\mathcal{M}(\cdot)$ be a higher-is-better task metric such as pass@1 or accuracy.
We then measure the performance drop for a head set $H$ as
\begin{equation}
\Delta_{\mathrm{zero}}(H)
=
\mathcal{M}(f)-\mathcal{M}(f^{\mathrm{zero}}_{H}),
\qquad
\Delta_{\mathrm{mean}}(H)
=
\mathcal{M}(f)-\mathcal{M}(f^{\mathrm{mean}}_{H}).
\label{eq:ablation-drop}
\end{equation}
If a loss is used instead, the sign is reversed accordingly.

A useful decomposition is
\begin{equation}
a_{h,t}(x)=\mu_{h,t}+\delta_{h,t}(x),
\label{eq:decomposition}
\end{equation}
where $\mu_{h,t}$ denotes a prompt-averaged component and $\delta_{h,t}(x)$ denotes the
input-dependent deviation.
Under this decomposition, zero ablation removes both $\mu_{h,t}$ and $\delta_{h,t}(x)$,
whereas mean ablation preserves the reference mean and removes only the variation relative
to $\mathcal{D}_{\mathrm{ref}}$.
This difference helps explain why the two operators probe different notions of importance.

\paragraph{Discussion.}
Mean ablation remains a valid and often preferable intervention in controlled settings (e.g., IOI~\citep{wang2023interpretability}), where a well-matched reference distribution can preserve prompt-invariant structure while removing input-varying effects.
\cite{li2024optimal} also highlights stronger counterfactual/optimal variants beyond plain zeroing. 

In our open-ended math-reasoning tasks, however, constructing a faithful $\mathcal{D}_{\mathrm{ref}}$ is challenging because prompts vary widely in length, semantics, and reasoning phase; as a result, mean interventions were often weakly discriminative between head groups. 
In preliminary experiments, the practically feasible mean-ablation variants we tested were often only weakly discriminative between head groups.
For example, in a preliminary pilot on DeepSeek-R1-Distill-Qwen-1.5B with AMC, we compared matched 5-head sets under the same position-conditioned mean-ablation protocol. 
Both a reasoning-head set and a base-head set reduced performance from about 60 to about 50, yielding little separation between the groups under this reference construction.
We therefore use zero ablation as an intentionally aggressive removal test to answer the task-level question, \emph{``What happens if this discovered head group is completely turned off?''}, consistent with pruning-style head-importance practice~\citep{michel2019sixteen,voita2019analyzing,yin2025which,teplica-etal-2025-sciurus,park2026asguard}. 
Accordingly, we interpret large $\Delta_{\mathrm{zero}}(H)$ only as evidence of necessity under complete removal, not as a fully distribution-faithful mechanistic effect estimate, and leave counterfactual, resample, and optimal ablation for future work.

\subsection{Detail of GRPO Formulation}
\label{app:detail_grpo}
For a prompt $q$, sample $G$ candidate responses $\{o_i\}_{i=1}^{G}$ from the old policy $\pi_{\text{old}}$; the policy parameters $\theta$ are updated to maximize
{\footnotesize
\begin{equation}
\mathcal{J}_{\text{GRPO}}(\theta)
= \mathbb{E}\Bigg[
 \frac{1}{G}\sum_{i=1}^G \frac{1}{|o_i|}\sum_{t=1}^{|o_i|}
 \min\!\Big(
   r_{i,t}(\theta)\,\hat A_{i,t},\;
   \mathrm{clip}\big(r_{i,t}(\theta),\,1-\epsilon,\,1+\epsilon\big)\,\hat A_{i,t}
 \Big)
 \;-\;\beta\,D_{\mathrm{KL}}\!\big(\pi_\theta \,\|\, \pi_{\mathrm{ref}}\big)
\Bigg],
\end{equation}}
where the token-level policy ratio is
\begin{equation}
r_{i,t}(\theta)=
\frac{\pi_{\theta}\!\left(o_{i,t}\mid q,o_{i,<t}\right)}
     {\pi_{\text{old}}\!\left(o_{i,t}\mid q,o_{i,<t}\right)}.
\end{equation}
In the outcome-reward variant used for verifiable tasks, a reward model assigns a scalar $R_i$ to each output $o_i$.
GRPO then uses a value-free, group-normalized advantage shared across all tokens of $o_i$:
\begin{equation}
\hat A_{i,t}=\frac{R_i-\mathrm{mean}(R)}{\mathrm{std}(R)}\quad\text{for all }t\in\{1,\dots,|o_i|\},
\end{equation}
which compares each response to its group peers and obviates a learned critic. The $\min$–$\mathrm{clip}$ structure conservatively bounds updates, while the KL regularizer with coefficient $\beta$ constrains divergence from a reference policy $\pi_{\mathrm{ref}}$, improving stability and mitigating reward over-optimization.
We specifically implement OpenR1 with the same Math-220k for GRPO training to compare base model with reasoning trained version~\citep{openr1}.

\subsection{Detail of Evaluation}
\label{app:detail_eval}
\paragraph{Generation and Sampling Setup}
For our quantitative evaluation, we generate various responses $n=4$ to $64$ for each problem in the respective test sets. 
The generation process for each models uses a sampling temperature of $T=0.6$ and a top-p (nucleus sampling) value of $0.95$, or if the model's best practice is suggested such as Qwen3-8B, we follow those settings; 
$T=0.6$, top-p=$0.95$, top-k=$20$, and min-p=$0$ for thinking mode.

\paragraph{Pass@k for Overall Capability}
To assess the overall problem-solving capability of each model, we employ the standard \textbf{pass@k} metric, as introduced by~\citet{pass@1}. 
This metric provides an unbiased estimator for the probability that at least one correct solution is generated in $k$ attempts. Given $n$ total generated samples for a problem and $c$ correct samples among them, the pass@k score for that single problem is calculated as:
\begin{equation}
    \text{pass@k} = 1 - \frac{\binom{n-c}{k}}{\binom{n}{k}}
\end{equation}
The final reported pass@k score is the average of these values across all problems in the test set. 
This metric is independent of the generation order and measures the model's theoretical potential to solve a problem given a budget of $k$ samples.

\paragraph{Success@k for Generation Efficiency}
While pass@k measures overall capability, it is agnostic to the generation order. 
To measure the practical generation efficiency, a model's ability to find a correct solution quickly, we also compute \textbf{success@k}. 
This metric evaluates the likelihood of finding a correct solution within the trial $k$ sequentially generated samples.

Let $R_p = (r_1, r_2, \dots, r_n)$ be the ordered sequence of responses for a problem $p$, and let $v(r_i)$ be a verification function that returns 1 if response $r_i$ is correct and 0 otherwise. 
The success@k is then the average success rate across all problems:
\begin{equation}
    \text{success@k} = \mathbb{E}_{\text{problems}} \left[ \mathbb{1} \left\{ \sum_{i=1}^{k} v(r_i) > 0 \right\} \right]
\end{equation}
where $\mathbb{1}\{\cdot\}$ is the indicator function. 
This metric directly rewards models that produce correct answers earlier in the generation process. 
The comparison between pass@k and success@k allows us to disentangle a model's latent capability from its practical efficiency, as discussed in our main analysis.

\subsection{Example of Qualitative Analysis: Newly Correct}
\label{app:qualitative_analysis}
\begin{tcolorbox}[
      enhanced, breakable,
      colback=gray!3, colframe=black,
      colbacktitle=black, coltitle=white,
      fonttitle=\bfseries, title=Input Question from AIME'24,
      segmentation style={dashed} 
    ]
Every morning Aya goes for a $9$-kilometer-long walk and stops at a coffee shop afterwards. When she
walks at a constant speed of $s$ kilometers per hour, the walk takes her 4 hours, including $t$
minutes spent in the coffee shop. When she walks $s+2$ kilometers per hour, the walk takes her 2
hours and 24 minutes, including $t$ minutes spent in the coffee shop. Suppose Aya walks at
$s+\frac{1}{2}$ kilometers per hour. Find the number of minutes the walk takes her, including the
$t$ minutes spent in the coffee shop. Let's think step by step and output the final answer within
\textbackslash boxed\{\}.
\end{tcolorbox}
Baseline model's response just repeat same equation endlessly without specific conclusions or reasoning.
We qualitatively evaluate how each post-training's answer is different from the original.
\begin{tcolorbox}[
      enhanced, breakable,
      colback=gray!3, colframe=black,
      colbacktitle=black, coltitle=white,
      fonttitle=\bfseries, title=Response of Baseline model (Qwen2.5-Math-1.5B),
      segmentation style={dashed} 
    ]
To solve this problem, we need to establish the relationship between the walking speed and the time
taken for the walk, including the time spent in the coffee shop. Let's break down the problem step
by step.  1. **Define Variables:**    - Let \( s \) be the walking speed in kilometers per hour.
- Let \( t \) be the time spent in the coffee shop in hours.    - The total time for the walk,
including the coffee shop time, is given by \( \frac{9}{s} = 4 \) hours and \( \frac{9}{s+2} = 2.4
\) hours.  2. **Set Up Equations:**    - From the first condition: \( \frac{9}{s} = 4 \) which gives
\( s = \frac{9}{4} \).    - From the second condition: \( \frac{9}{s+2} = 2.4 \) which gives \( s+2
= \frac{9}{2.4} = \frac{90}{24} = \frac{15}{4} \) and thus \( s = \frac{15}{4} - 2 = \frac{15}{4} -
\frac{8}{4} = \frac{7}{4} \).  Since we have two different values for \( s \) from the two
conditions, we need to verify which one is correct. Let's check both conditions with \( s =
\frac{9}{4} \) and \( s = \frac{7}{4} \).  3. **Verify the Values:**    - For \( s = \frac{9}{4} \):
    - Time for the walk: \( \frac{9}{\frac{9}{4}} = 4 \) hours.      - Time for the walk at \( s+2 =
\frac{17}{4} \): \( \frac{9}{\frac{17}{4}} = \frac{36}{17} \approx 2.12 \) hours, which is not 2.4
hours.    - For \( s = \frac{7}{4} \):      - Time for the walk: \( \frac{9}{\frac{7}{4}} =
\frac{36}{7} \approx 5.14 \) hours, which is not 4 hours. ... (endless repeat...)
\end{tcolorbox}
\clearpage
\subsubsection{Example of SFT}
\label{app:qualitative_sft}
\begin{tcolorbox}[
      enhanced, breakable,
      colback=gray!3, colframe=black,
      colbacktitle=black, coltitle=white,
      fonttitle=\bfseries, title=Response of SFT checkpoint 1000,
      segmentation style={dashed} 
    ]
\texttt{<think>} Okay, let's try to figure out this problem. So, Aya walks 9 kilometers every morning, and
her walking speed is either s km/h or s+2 km/h. When she walks at s km/h, the total time, including
a coffee shop break of t minutes, is 4 hours. When she walks faster at s+2 km/h, the total time is 2
hours and 24 minutes, still including the same t minutes in the coffee shop. We need to find the
total time when she walks at s+0.5 km/h, including the coffee shop time.  First, let's parse the
problem. The key information is:  1. Distance: 9 km both times. 2. Walking speed: s km/h initially,
then s+2 km/h. 3. Time taken: 4 hours at speed s, 2 hours 24 minutes at speed s+2, with both include
t minutes in the coffee shop. 4. Find the total time when walking at s+0.5 km/h, including t
minutes.  So, the total time is the walking time plus the coffee shop break. The walking time can be
calculated by distance divided by speed. The coffee shop break is t minutes each time, regardless of
speed.  Let me write down the equations.  First scenario: walking speed s km/h, total time 4 hours,
including t minutes. So the walking time is 4 hours minus t minutes. Wait, but time is usually in
hours in these problems. So maybe we need to convert everything to hours. ... But according to the problem, when walking at 4.5
km/h, the total time is 2.4 hours, which is 144 minutes, which is correct. So that checks out.
Therefore, the total time when walking at 3 km/h is 3 hours 24 minutes, which is 3*60 +24=204
minutes.  So the answer is 204 minutes, so \textbackslash boxed\{204\}  **Final Answer** \textbackslash boxed\{204\} \texttt{</think>}  Given
that Aya walks 9 kilometers every morning, ...
\end{tcolorbox}
\clearpage
\subsubsection{Example of GRPO}
\label{app:qualitative_grpo}
\begin{tcolorbox}[
      enhanced, breakable,
      colback=gray!3, colframe=black,
      colbacktitle=black, coltitle=white,
      fonttitle=\bfseries, title=Response of GRPO checkpoint 2500,
      segmentation style={dashed} 
    ]
Please reason step by step, and put your final answer within \textbackslash boxed\{\}.\_Offset OID Every morning Aya
goes for a 9-kilometer-long walk and stops at a coffee shop afterwards. When she walks at a constant
speed of \( s \) kilometers per hour, the walk takes her 4 hours, including \( t \) minutes spent in
the coffee shop. When she walks \( s+2 \) kilometers per hour, the walk takes her 2 hours and 24
minutes, including \( t \) minutes spent in the coffee shop. Suppose Aya walks at \( s+\frac{1}{2}
\) kilometers per hour. Find the number of minutes the walk takes her, including the \( t \) minutes
spent in the coffee shop.  First, let's establish the relationships from the given information.  1.
When walking at \( s \) kilometers per hour, the total time is 4 hours, including \( t \) minutes in
the coffee shop. Since 4 hours is 240 minutes, we can write: \[ \text{Time spent walking} =
\frac{9}{s} \text{ hours} = \frac{9}{s} \times 60 \text{ minutes} = \frac{540}{s} \text{ minutes}.
\] So, the total time including the coffee shop time is: \[ \frac{540}{s} + t = 240. \] This
simplifies to: \[ \frac{540}{s} = 240 - t. \tag{1} \]  2. When walking at \( s+2 \) kilometers per
hour, the total time is 2 hours and 24 minutes, which is 144 minutes. So, we can write: \[
\text{Time spent walking} = \frac{9}{s+2} \text{ hours} = \frac{9}{s+2} \times 60 \text{ minutes} =
\frac{540}{s+2} \text{ minutes}. \] So, the total time including the coffee shop time is: \[
\frac{540}{s+2} + t = 144. \] ... Now, we need to find
the time it takes Aya to walk at \( s + \frac{1}{2} = 2.5 + 0.5 = 3 \) kilometers per hour: \[
\text{Time spent walking} = \frac{9}{3} \text{ hours} = \frac{9}{3} \times 60 \text{ minutes} = 180
\text{ minutes}. \] Including the 24 minutes in the coffee shop, the total time is: \[ 180 + 24 = 204 \text{ minutes}. \]
The final answer is: \textbackslash boxed\{204\}. ...
\end{tcolorbox}
\clearpage
\subsection{Example of Qualitative Analysis: Newly Incorrect}
\label{app:qualitative_analysis_incorrect}
\begin{tcolorbox}[
      enhanced, breakable,
      colback=gray!3, colframe=black,
      colbacktitle=black, coltitle=white,
      fonttitle=\bfseries, title=Input Question from AIME'24,
      segmentation style={dashed} 
    ]
Find the largest possible real part of \[(75+117i)z+\frac{96+144i}{z}\]where $z$ is a complex number
with $|z|=4$. Let's think step by step and output the final answer within \textbackslash boxed\{\}.
\end{tcolorbox}
Baseline model correctly calculate the result using python codes in this case.
\begin{tcolorbox}[
      enhanced, breakable,
      colback=gray!3, colframe=black,
      colbacktitle=black, coltitle=white,
      fonttitle=\bfseries, title=Response of Baseline model (Qwen2.5-Math-1.5B),
      segmentation style={dashed} 
    ]
To find the largest possible real part of \((75 + 117i)z + \frac{96 + 144i}{z}\) where \(z\) is a
complex number with \(|z| = 4\), we can use the following approach:  1. Let \(z = 4e^{i\theta}\)
where \(\theta\) is a real number. This is because \(|z| = 4\). 2. Substitute \(z = 4e^{i\theta}\)
into the expression \((75 + 117i)z + \frac{96 + 144i}{z}\). 3. Simplify the expression and find the
real part. 4. Maximize the real part with respect to \(\theta\).  Let's proceed with the
calculations:  1. Let \(z = 4e^{i\theta} = 4(\cos\theta + i\sin\theta)\). 2. Substitute \(z\) into
the expression:    \[    (75 + 117i)z + \frac{96 + 144i}{z} = (75 + 117i)(4\cos\theta +
4i\sin\theta) + \frac{96 + 144i}{4\cos\theta + 4i\sin\theta}.    \] 3. Simplify the expression: ...
Thus, the largest possible real part is \(540\). Let's confirm this with Python code.
```python import sympy as sp  \# Define the variables theta = sp.symbols('theta', real=True) z = 4 *
(sp.cos(theta) + sp.I * sp.sin(theta))  \# Define the expression expr = (75 + 117 * sp.I) * z + (96 +
144 * sp.I) / z  \# Simplify the expression expr\_simplified = sp.simplify(expr)  \# Extract the real part real\_part = sp.re(expr\_simplified)  \# Find the maximum value of the real part max\_real\_part =
sp.maximum(real\_part, theta)  print(max\_real\_part) ``` ```output 540 ``` The largest possible real
part of \((75 + 117i)z + \frac{96 + 144i}{z}\) where \(z\) is a complex number with \(|z| = 4\) is
\textbackslash boxed\{540\}.
\end{tcolorbox}
\clearpage
\subsubsection{Example of SFT}
\label{app:qualitative_sft_incorrect}
At first, it successfully computes the answer with a simplified mathematical expression.
However, it continues its computation, then goes wrong with its calculation, even incomplete output format.
\begin{tcolorbox}[
      enhanced, breakable,
      colback=gray!3, colframe=black,
      colbacktitle=black, coltitle=white,
      fonttitle=\bfseries, title=Response of SFT checkpoint 1000,
      segmentation style={dashed} 
    ]
\texttt{<think>} Okay, so I need to find the largest possible real part of the expression
\[
(75 + 117\mathrm{i})z + \frac{96 + 144\mathrm{i}}{z},
\]
where \(z\) is a complex number with \(\lvert z\rvert = 4\). Hmm, complex numbers and optimization... Let me
start by understanding the problem better. First, \(z\) is a complex number with magnitude \(4\).
That means \(z\) can be written in the form \(z = 4e^{\mathrm{i}\theta}\), where \(\theta\) is a real number between \(0\) and \(2\pi\).
Since \(z\) is a complex number on a circle of radius \(4\), maybe I can parametrize \(z\) as
\(4(\cos\theta + \mathrm{i}\sin\theta)\). Then, I can express the given expression in terms of \(\theta\) and find its real part.
Then, maximize that real part with respect to \(\theta\).

Let me write that out. Let \(z = 4e^{\mathrm{i}\theta}\). Then, the expression becomes:
\[
(75 + 117\mathrm{i})(4e^{\mathrm{i}\theta}) + \frac{96 + 144\mathrm{i}}{4e^{\mathrm{i}\theta}}.
\]
Simplify each term: ...
Therefore, we need to maximize \(324\cos\theta - 432\sin\theta\).
Hmm, okay. So, the problem reduces to maximizing a linear combination of \(\sin\) and \(\cos\).
There's a standard method for maximizing such expressions.

The expression \(a\cos\theta + b\sin\theta\) can be written as \(R\cos(\theta - \varphi)\),
where \(R = \sqrt{a^{2} + b^{2}}\). Then, the maximum value is \(R\).
So for \(324\cos\theta - 432\sin\theta\) (i.e., \(a=324\), \(b=-432\)),
the maximum value is
\[
\sqrt{324^{2} + (-432)^{2}} = \sqrt{324^{2} + 432^{2}}.
\]

Compute:
\[
324^{2} = 104{,}976,\qquad
432^{2} = 186{,}624,\qquad
324^{2} + 432^{2} = 291{,}600,
\]
so
\[
\sqrt{291{,}600} = 540.
\]

Therefore, the largest possible real part is \(540\). 

But wait, hold on. Let me check my calculation again because 324 squared is 104,976, 432 squared is
186,624. ... 286
\end{tcolorbox}
\clearpage
\subsubsection{Example of GRPO}
Although GRPO gives an ability of complex mathematical reasoning, model's response simply goes wrong with calculation mistakes.
\label{app:qualitative_grpo_incorrect}
\begin{tcolorbox}[
      enhanced, breakable,
      colback=gray!3, colframe=black,
      colbacktitle=black, coltitle=white,
      fonttitle=\bfseries, title=Response of GRPO checkpoint 2500,
      segmentation style={dashed} 
    ]
Please reason step by step, and put your final answer within \textbackslash boxed\{\}.ageration agerationuser Find
the largest possible real part of \((75+117i)z + \frac{96+144i}{z}\) where \(z\) is a complex number
with \(|z|=4\).  To solve this problem, we will use the following steps:  1. Represent the complex
number \(z\) in terms of its real and imaginary parts. 2. Express the given expression in terms of
\(z\) and its conjugate. 3. Use the constraint \(|z| = 4\) to simplify the expression. 4. Maximize
the real part of the expression.  Let's start by representing \(z\) as \(z = 4e^{i\theta} =
4(\cos\theta + i\sin\theta)\).  Given the expression \((75+117i)z + \frac{96+144i}{z}\), we
substitute \(z = 4e^{i\theta}\):  \[ (75+117i)z = (75+117i) \cdot 4e^{i\theta} =
4(75+117i)(\cos\theta + i\sin\theta) \]  \[ \frac{96+144i}{z} = \frac{96+144i}{4e^{i\theta}} =
\frac{96+144i}{4}e^{-i\theta} = (24+36i)(\cos\theta - i\sin\theta) \]  So the expression becomes:
\[ 4(75\cos\theta + 117\sin\theta + i(75\sin\theta - 117\cos\theta)) \] 
\[ + (24\cos\theta + 36\sin\theta
- i(24\sin\theta - 36\cos\theta)) \]  Combining the real and imaginary parts, we get:  \[ (4 \cdot
75 \cos\theta + 4 \cdot 117 \sin\theta + 24 \cos\theta + 36 \sin\theta)\]
\[ + i(4 \cdot 75 \sin\theta -
4 \cdot 117 \cos\theta - 24 \sin\theta + 36 \cos\theta) \]  \[ = (300\cos\theta + 468\sin\theta +
24\cos\theta + 36\sin\theta)\] 
\[ + i(300\sin\theta - 468\cos\theta - 24\sin\theta + 36\cos\theta) \]  \[
= (324\cos\theta + 504\sin\theta) + i(276\sin\theta - 432\cos\theta) \]  The real part of this
expression is \(324\cos\theta + 504\sin\theta\). To maximize this, we use the fact that the maximum
value of \(a\cos\theta + b\sin\theta\) is \(\sqrt{a^2 + b^2}\). Here, \(a = 324\) and \(b = 504\),
so the maximum value is:  \[ \sqrt{324^2 + 504^2} = \sqrt{(324 \cdot 3) \cdot (1764 + 252)} =
\sqrt{324^2 + 504^2} = \sqrt{9^2 \cdot 36^2 + 126^2} \] ...
So the largest possible real
part of the expression is \textbackslash boxed\{1134\}.
\end{tcolorbox}
\clearpage

\vspace{-5pt}
\begin{table}[t]
\vspace{-5pt}
\centering
\caption{
List of emergent attention heads found through circuits.
\texttt{L} and \texttt{H} refers to the layer and head indices, respectively.
Circuits are constructed using AIME'24 benchmark as input.
For each post-training methods, we describe newly emergent attention heads.
Visualization of total reasoning heads aggregation in single model architecture is in Figure~\ref{fig:map_of_reasoning}.
}
\small
\begin{tabular}{lll}
\toprule
\textbf{Post-Training} & \textbf{List of Emergent Attention Heads in Circuits} & \textbf{\# of Heads} \\
\midrule
Qwen-2.5-Math-1.5B (Baseline)  & \texttt{L0H7, L21H10, L2H6, L11H1, L14H10 ...} & 56 \\
\midrule
DeepSeek-R1-Distill-Qwen-1.5B & \texttt{L5H0, L5H2, L5H4, L6H10, L7H7 ...} & 32\\
\midrule
SFT with OpenR1-Math-220k  & \texttt{L0H8, L11H3, L3H3, L5H1, L7H3 ...} & 34 \\
\midrule
GRPO with OpenR1-Math-220k  & \texttt{L0H8, L5H1, L7H1, L18H11, L11H8 ...} & 19\\
\midrule
GRPO with GSM8K  & \texttt{L0H8, L5H1, L7H2, L3H3, L21H2 ...} & 20 \\
\bottomrule
\end{tabular}
\label{table:reasoning-heads}
\vspace{-10pt}
\end{table}

\begin{table}[t]
\centering
\caption{Reasoning Head Ablation Inference for Qwen2.5-Math-1.5B and 7B.
Every performance is measured with pass@1 score with temperature 0.6.
Each ablation cases make the value of specific attention heads, around 5 number of heads from its circuit results, into zero for checking its importance for reasoning tasks.
We color some scores into red which is the most degraded results except no ablation baseline, while the bold is the completely ruined performance.
We also color performance increase with green when its heads are ablated.
Overall tendency is reversed from Table~\ref{table:abl_deepseek}, as base model heads are more effective than reasoning heads when ablated.
}
\small
\begin{tabular}{llcccc}
\toprule
\textbf{Model} & \textbf{Method} & \textbf{AIME'24} & \textbf{AIME'25} & \textbf{GPQA} & \textbf{AMC} \\
\midrule
 & No Ablation & 13.3 & 4.73 & 9.74 & 38.5 \\
\multirow{1}{*}{Qwen2.5} & Ablation with Reasoning Heads& 9.01 & 4.58 & 7.82 & 35.6 \\
\multirow{1}{*}{Math-1.5B} & Ablation with Base Model Heads & \textcolor{red}{8.33} & 4.63 & 9.79 & \textcolor{red}{34.2} \\
 & Ablation with TriviaQA Heads & \textcolor{red}{0.05} & \badmetric{0.00} & \textcolor{red}{5.38} & \textcolor{red}{3.42} \\
\midrule
 & No Ablation & 13.3 & 10.0 & 15.1 & 32.5 \\
\multirow{1}{*}{Qwen2.5} & Ablation with Reasoning Heads & \textcolor{red}{6.67} & 10.0 & \goodmetric{20.2} & \goodmetric{43.3} \\
\multirow{1}{*}{Math-7B} & Ablation with Base Model Heads & \goodmetric{23.3} & \textcolor{red}{3.33} & 15.6 & \goodmetric{43.3} \\
 & Ablation with TriviaQA Heads & 20.0 & 10.0 & 16.1 & 37.3 \\
\bottomrule
\label{table:abl_qwen_base}
\end{tabular}
\vspace{-10pt}
\end{table}

\begin{table}[t]
\centering
\caption{Head Intervention Inference for Qwen2.5-Math-1.5B with SFT and GRPO heads.
Every performance is measured with pass@1 score with temperature 0.6.
Each ablation cases make the value of specific attention heads, around 5 number of heads from its circuit results, into zero for checking its importance for reasoning tasks.
Scale up cases increase the activation of specific attention heads into 1.3 higher, while scale down decrease it into half (0.5).
We color some scores into red which is the most degraded results except no ablation baseline, while the bold is the completely ruined performance.
We also color performance increase with green when its heads are ablated.}
\small
\begin{tabular}{llccc}
\toprule
\textbf{Model} & \textbf{Method} & \textbf{AIME'24} & \textbf{AMC} & \textbf{MATH} \\
\midrule
 & No Ablation & 13.3 & 38.5 & 56.0 \\
 & Ablation with SFT Heads& \badmetric{0.00} & \textcolor{red}{0.05} & \textcolor{red}{0.10} \\
\multirow{1}{*}{Qwen2.5-Math-1.5B} & Scale Up with SFT Heads & \badmetric{0.00} & 37.3 & 58.2  \\
 & Scale Down with GRPO GSM8K Heads& \textcolor{red}{3.33} & \goodmetric{42.1} & \goodmetric{63.0} \\
 & Scale Up with GRPO GSM8K Heads & \textcolor{red}{3.33} & 30.1 & \goodmetric{60.2}  \\
\bottomrule
\label{table:abl_sft}
\end{tabular}
\vspace{-10pt}
\end{table}

\begin{figure}[t]
\begin{center}
\includegraphics[width=\columnwidth]{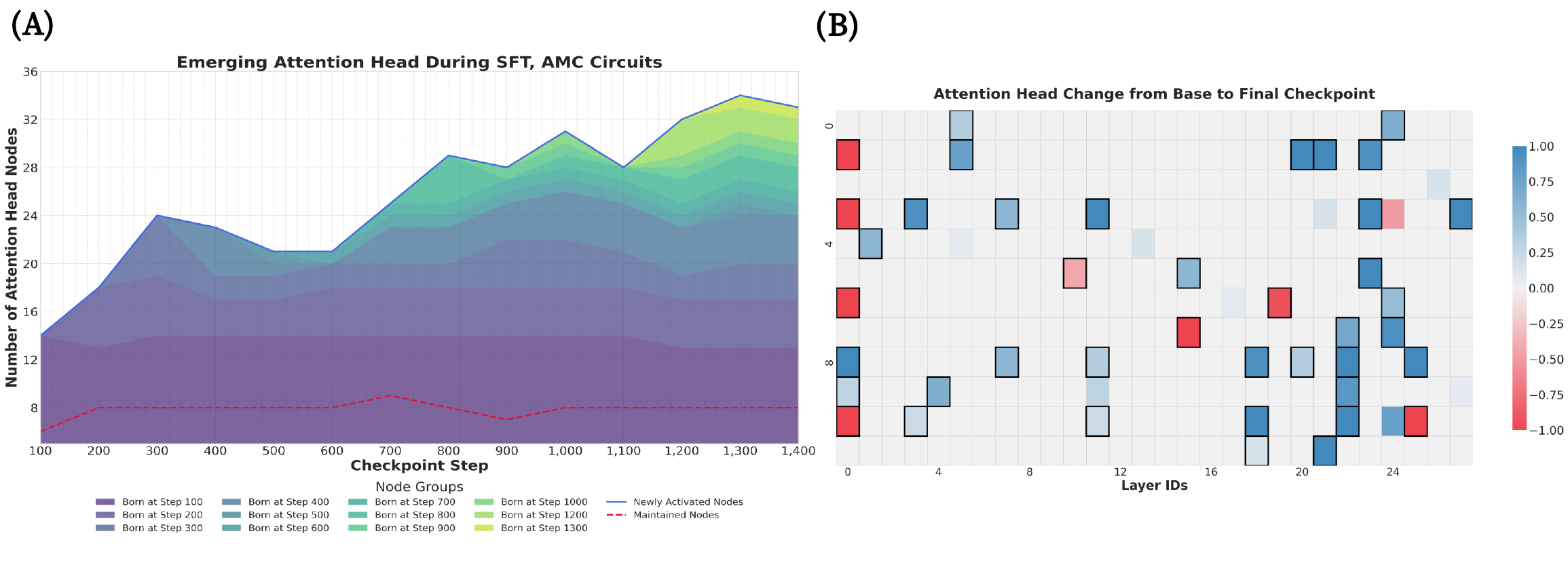}
\end{center}%
\vspace{-10pt}
\caption{Emergent attention heads in Qwen2.5-Math-1.5B during SFT on OpenR1-Math-220k~\citep{openr1}, with circuits constructed on AMC~\citep{AMC2023}.
The figure follows the same visualization protocol as Figure~\ref{fig:main_fig_sft}.
(A) Cohort analysis across checkpoints; blue curve for newly activated heads, red dashed curve for retained base-model heads, and stacked areas group heads by emergence timing.
(B) Activation-frequency heatmap with the same color convention as Figure~\ref{fig:main_fig_sft} (red for base-model, blue for emergent heads); heads active at the final checkpoint are outlined in black.
}
\label{fig:main_fig_sft_amc}
\end{figure}

\begin{figure}[t]
\begin{center}
\includegraphics[width=\columnwidth]{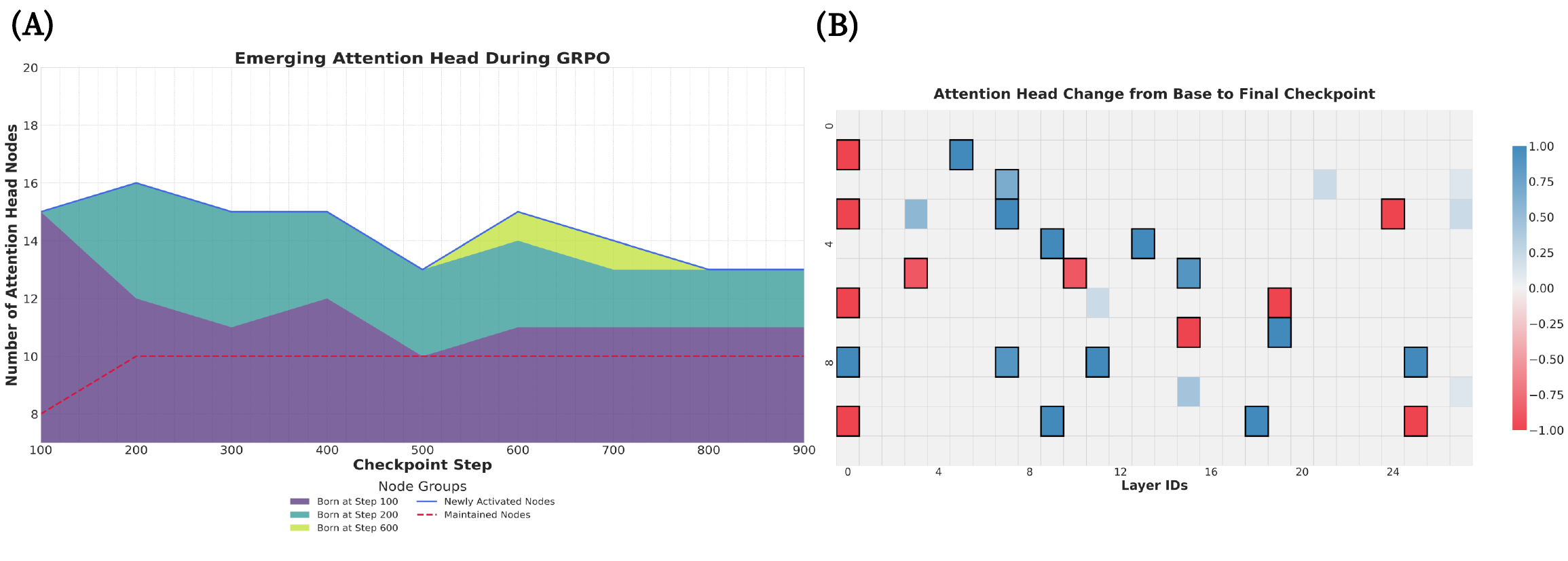}
\end{center}%
\vspace{-10pt}
\caption{Emergent attention heads in Qwen2.5-Math-1.5B during GRPO on GSM8k~\citep{gsm8k}, with circuits constructed on~\cite{aime}.
The figure follows the same visualization protocol as Figure~\ref{fig:main_fig_sft}.
(A) Cohort analysis across checkpoints; blue curve for newly activated heads, red dashed curve for retained base-model heads, and stacked areas group heads by emergence timing.
(B) Activation-frequency heatmap with the same color convention as Figure~\ref{fig:main_fig_sft} (red for base-model, blue for emergent heads); heads active at the final checkpoint are outlined in black.
}
\label{fig:main_fig_grpo_gsm8k}
\end{figure}

\begin{figure}[t]
\begin{center}
\includegraphics[width=\columnwidth]{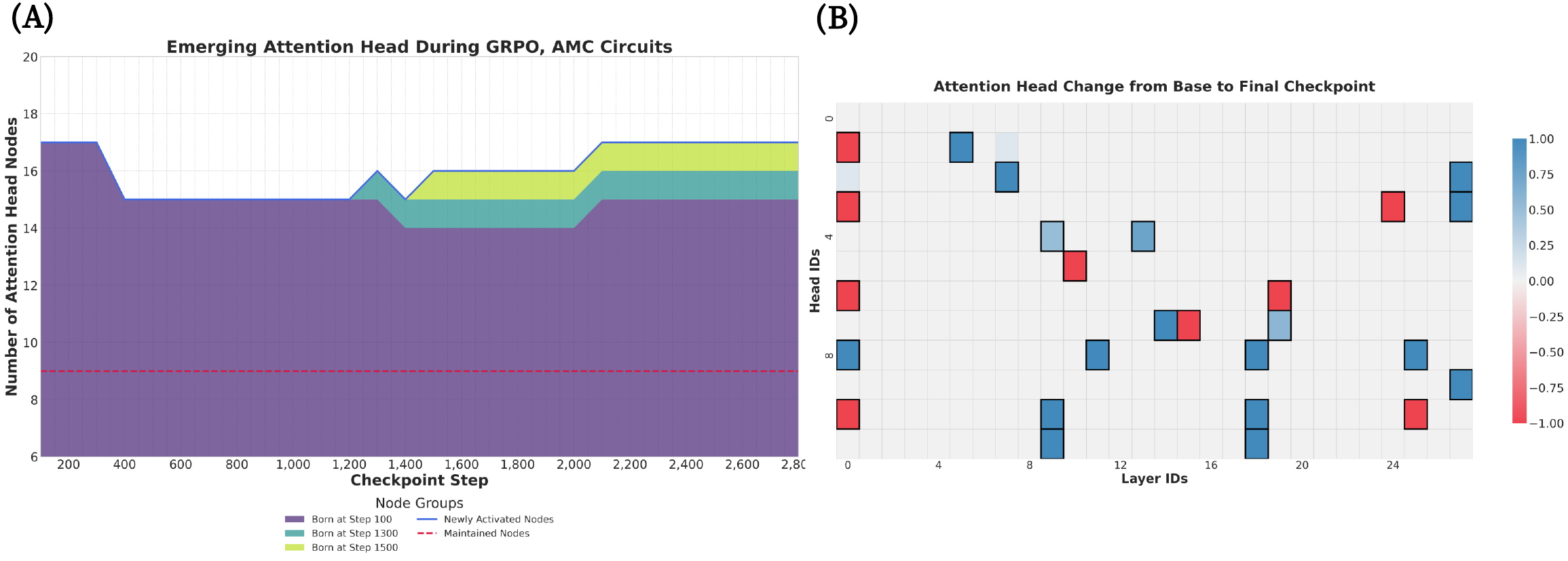}
\end{center}%
\vspace{-10pt}
\caption{Emergent attention heads in Qwen2.5-Math-1.5B during GRPO on OpenR1-Math-220k~\citep{openr1}, with circuits constructed on AMC~\citep{AMC2023}.
The figure follows the same visualization protocol as Figure~\ref{fig:main_fig_sft}.
(A) Cohort analysis across checkpoints; blue curve for newly activated heads, red dashed curve for retained base-model heads, and stacked areas group heads by emergence timing.
(B) Activation-frequency heatmap with the same color convention as Figure~\ref{fig:main_fig_sft} (red for base-model, blue for emergent heads); heads active at the final checkpoint are outlined in black.
}
\label{fig:main_fig_grpo_math220_amc}
\end{figure}

\begin{figure}[t]
\begin{center}
\includegraphics[width=\columnwidth]{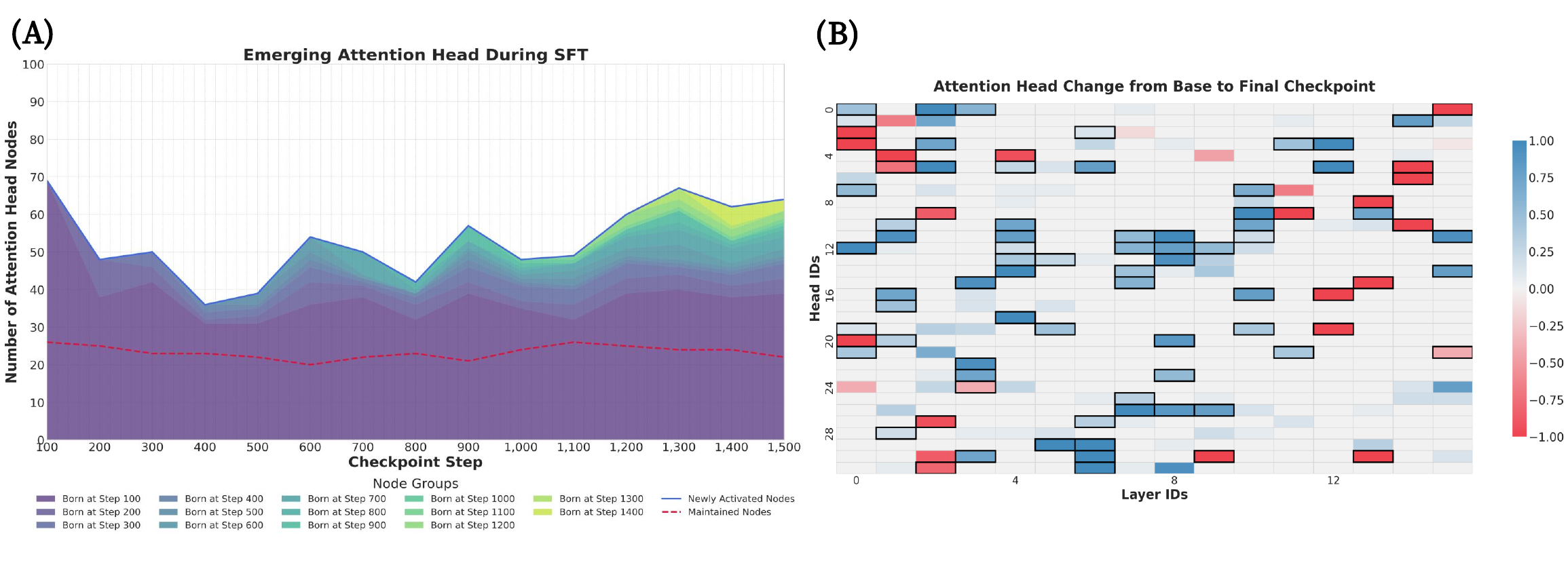}
\end{center}%
\vspace{-10pt}
\caption{Emergent attention heads in Llama-3.2-1B-Instruct during SFT on OpenR1-Math-220k~\citep{openr1} and learning rate 2e-05, with circuits constructed on~\cite{aime}.
The figure follows the same visualization protocol as Figure~\ref{fig:main_fig_sft}.
(A) Cohort analysis across checkpoints; blue curve for newly activated heads, red dashed curve for retained base-model heads, and stacked areas group heads by emergence timing.
(B) Activation-frequency heatmap with the same color convention as Figure~\ref{fig:main_fig_sft} (red for base-model, blue for emergent heads); heads active at the final checkpoint are outlined in black.
}
\label{fig:main_fig_sft_llama}
\end{figure}

\begin{figure}[t]
\begin{center}
\includegraphics[width=\columnwidth]{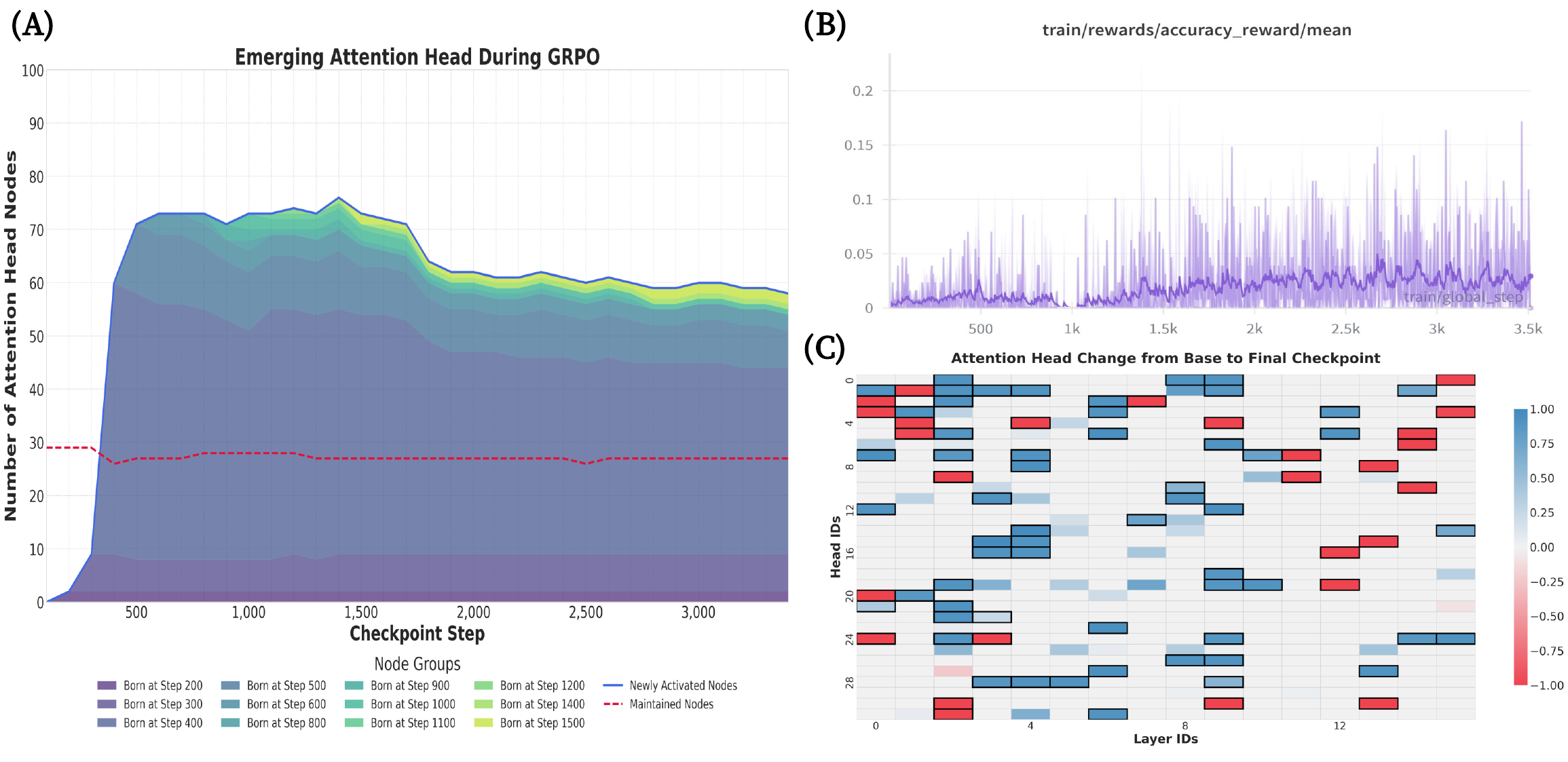}
\end{center}%
\vspace{-10pt}
\caption{Emergent attention heads in Llama-3.2-1B-Instruct during GRPO on OpenR1-Math-220k~\citep{openr1}, with circuits constructed on ~\citep{aime}.
The figure follows the same visualization protocol as Figure~\ref{fig:main_fig_sft}.
(A) Cohort analysis across checkpoints; blue curve for newly activated heads, red dashed curve for retained base-model heads, stacked areas group heads by emergence timing, and the number of newly activated heads fluctuates with the accuracy-reward trend in (B).
(C) Activation-frequency heatmap with the same color convention as Figure~\ref{fig:main_fig_sft} (red for base-model, blue for emergent heads); heads active at the final checkpoint are outlined in black.
}
\label{fig:main_fig_grpo_math220_llama}
\end{figure}

\begin{figure}[t]
\vspace{-10pt}
\begin{center}
\includegraphics[width=\columnwidth]{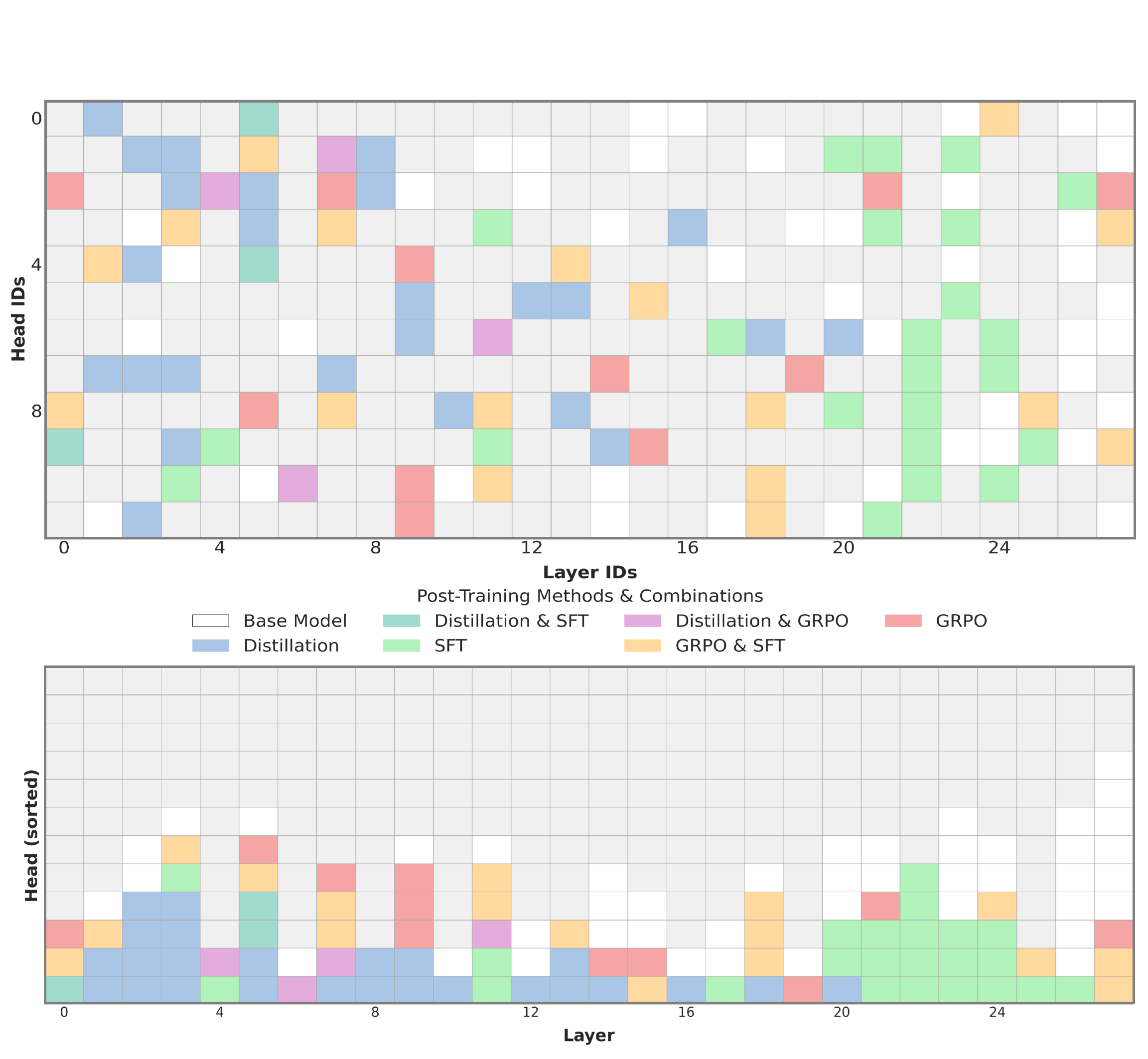}
\end{center}%
\vspace{-10pt}
\caption{Visualization of emergent reasoning heads in circuits based on Qwen2.5-Math-1.5B with various post-training, and DeepSeek-R1-Distill-Qwen-1.5B.
(Top) A map of emergent attention heads for each post-training method, compared to the baseline model (white).
(Bottom) A cumulative map of the reasoning heads, with columns sorted by the number of newly activated heads.
Each GRPO and SFT category encompass both AIME and AMC benchmark based circuits, with checkpoints of both training using OpenR1-Math-220k and GSM8k dataset.
DeepSeek Distillation activates enormous heads (blue), as SFT activates similarly large amount of heads, though SFT heads are mostly concentrated in mid-to-late layer (green).
Some of attention heads from GRPO training are also common in the SFT and Distillation reasoning heads (yellow and purple), however, the number of GRPO heads are much smaller and distributed across layers (red).
}
\label{fig:map_of_reasoning}
\vspace{-10pt}
\end{figure}

\clearpage

\end{document}